%% file: egpaper_final.tex
\documentclass[10pt,twocolumn,letterpaper]{article}

\usepackage{iccv}
\usepackage{times}
\usepackage{epsfig}
\usepackage{graphicx}
\usepackage{amsmath}
\usepackage{amssymb}

\usepackage{booktabs}

\usepackage[accsupp]{axessibility}  % Improves PDF readability for those with disabilities.
\usepackage{times}
\usepackage{epsfig}
\usepackage{enumitem} % for itemized bullets
\usepackage{euscript} % for curly N
\usepackage{diagbox} % for diagonal line in a table
\usepackage{arydshln} % for dashed line in a table
\usepackage{multirow} % for merging cells in a table
\usepackage[hang,flushmargin]{footmisc} 

\usepackage[pagebackref=true,breaklinks=true,colorlinks,bookmarks=false]{hyperref}
\iccvfinalcopy % *** Uncomment this line for the final submission

\def\iccvPaperID{10225} % *** Enter the ICCV Paper ID here

\ificcvfinal\fi

\makeatletter
\newcommand{\printfnsymbol}[1]{%
  \textsuperscript{\@fnsymbol{#1}}%
}
\makeatother

\input{macros.tex}

\mathchardef\mhyphen="2D

\DeclareMathOperator*{\argmin}{\arg\!\min}

\begin{document}

%%%%%%%%% TITLE
\title{SAGA: Spectral Adversarial Geometric Attack on 3D Meshes}

\author{Tomer Stolik\printfnsymbol{1}\\
Tel Aviv University\\
{\tt\small tomerstolik@mail.tau.ac.il}
\and
Itai Lang\printfnsymbol{1}\\
Tel Aviv University\\
{\tt\small itailang@mail.tau.ac.il}
\and
Shai Avidan\\
Tel Aviv University\\
%First line of institution3 address\\
{\tt\small avidan@eng.tau.ac.il}
}

\maketitle
\ificcvfinal\thispagestyle{empty}\fi

%-------------------------------------------------------------------------
%%%%%%%%% ABSTRACT
\input{00_abstract.tex}

%-------------------------------------------------------------------------
%%%%%%%%% INTRODUCTION
\input{01_introduction.tex}

%-------------------------------------------------------------------------
%%%%%%%%% Related Work
\input{02_related_work.tex}

%-------------------------------------------------------------------------
%%%%%%%%% METHOD
\input{03_method.tex}
%\input{poster/equations}

%-------------------------------------------------------------------------
%%%%%%%%% RESULTS
\input{04_results.tex}

%-------------------------------------------------------------------------
%%%%%%%%% CONCLUSIONS
\input{05_conclusions.tex}

%-------------------------------------------------------------------------
%%%%%%%%% REFERENCES
{\small
\bibliographystyle{ieee_fullname}
\bibliography{egbib}
}

%-------------------------------------------------------------------------
%%%%%%%%% SUPPLEMENTARY MATERIAL
\input{supplementary/supplementary.tex}

%-------------------------------------------------------------------------

\end{document}

%% file: macros.tex
\long\def\ignorethis#1{}

\usepackage{color}
\definecolor{gray}{rgb}{0.35,0.35,0.35}
\definecolor{red}{rgb}{1,0,0}
\definecolor{dark-green}{rgb}{0,0.4,0}
\definecolor{blue}{rgb}{0,0,1}
\definecolor{orange}{rgb}{1,0.55,0}
\definecolor{white}{rgb}{1,1,1}
\definecolor{black}{rgb}{0,0,0}
\definecolor{dark-brown}{rgb}{0.2,0.1,0}
\definecolor{light-blue}{rgb}{0.4,0.6,0.99}
\definecolor{dark-red}{rgb}{0.6,0,0}
\definecolor{light-red}{rgb}{1,0.2,0.6}
\definecolor{pink}{rgb}{1,0.2,0.6}
\definecolor{dark-pink}{rgb}{0.6,0,0.3}

\newcommand{\itai}[1]{{\color{black}#1}\normalfont}
\newcommand{\tomer}[1]{{\color{black}#1}\normalfont}

\newcommand{\newtxt}[1]{{\color{black}#1}\normalfont}

\newbox\jsavebox

%% file: 00_abstract.tex
\begin{abstract}
A triangular mesh is one of the most popular 3D data representations. As such, the deployment of deep neural networks for mesh processing is widely spread and is increasingly attracting more attention. However, neural networks are prone to adversarial attacks, where carefully crafted inputs impair the model's functionality. The need to explore these vulnerabilities is a fundamental factor in the future development of 3D-based applications. Recently, mesh attacks were studied on the semantic level, where classifiers are misled to produce wrong predictions. Nevertheless, mesh surfaces possess complex geometric attributes beyond their semantic meaning, and their analysis often includes the need to encode and reconstruct the geometry of the shape.

We propose a novel framework for a geometric adversarial attack on a 3D mesh autoencoder. In this setting, an adversarial input mesh deceives the autoencoder by forcing it to reconstruct a different geometric shape at its output. The malicious input is produced by perturbing a clean shape in the spectral domain. Our method leverages the spectral decomposition of the mesh along with additional mesh-related properties to obtain visually credible results that consider the delicacy of surface distortions\footnote{\url{https://github.com/StolikTomer/SAGA} \\
\printfnsymbol{1}Equal contribution}.

\end{abstract}

%% file: 01_introduction.tex
\section{Introduction} \label{sec:introduction}

\input{figures/teaser/teaser_pdf.tex}

\newtxt{A triangular mesh is the primary representation of 3D shapes, with applications in many safety-critical realms. In the medical field, incorrect perception of the geometric subtleties of an organ can lead to life-threatening errors. In robotics and automotive, a precise understanding of the geometry of obstacles is essential to prevent accidents. The security of facial modeling is also dependent on the accuracy of the processed geometry of the mesh.

Autoencoders (AEs) are one of the most prominent deep-learning tools to process the mesh's geometry. They are designed to capture geometric features which enable dimensionality reduction for both storage and communication purposes~\cite{gao2019sdm, chen2021learning}. Mesh AEs are also used for segmentation, self-supervised learning, and denoising tasks~\cite{mehr2019disconet, nousias2020fast, george20183d}.

Despite their tremendous achievements, neural networks are often found vulnerable to adversarial attacks. These attacks craft inputs that impair the victim network's behavior. Adversarial attacks were extensively studied in recent years, focusing especially on the \textit{semantic} level, where the input to a classifier is carefully modified in an imperceivable manner to mislead the network to an incorrect prediction. Semantic adversarial attacks are abundant in the case of 2D images~\cite{goodfellow2015explaining, papernot2016the, carlini2017towards}, and recently, semantic attacks on 3D representations have also drawn much attention, both on point clouds~\cite{xiang2019generating, hamdi2020advpc, wen2020geometry} and meshes~\cite{xiao2019meshadv, mariani2020generating, rampini2021universal, belder2022random}.

Nonetheless, the vulnerabilities of networks that process geometric attributes, such as AEs, have not been thoroughly investigated. AEs may be imperative to many practical mesh deployments and their credibility and robustness depend on the study of \textit{geometric} adversarial attacks.}

We propose a framework of a geometric adversarial attack on 3D meshes. Our attack, named SAGA, is \newtxt{exemplified} in Figure~\ref{fig:teaser}. The input mesh of the sphere is perturbed and fed into an AE that reconstructs a \textit{geometrically different} output\newtxt{, \ie, a cube!} Ideally, the deformation of the input should be unapparent and yet effectively modify the output geometry.

In our attack, we aim to reconstruct the geometry of a \textit{specific target} mesh by perturbing a clean \textit{source} mesh into a malicious input. We present a white-box setting, where we have access to the AE and we optimize the attack according to its output. A black-box framework is also explored by transferring the adversarial examples to other unseen AEs.

Mesh perturbations include shifts of vertices that affect their adjacent edges and faces and possibly result in noticeable topological disorders, such as self-intersections. Therefore, concealed perturbations must address the inherent topological constraints of the mesh. To cope with the fragility of the mesh surface, we apply the perturbations in the spectral domain defined by the eigenvectors of the Laplace-Beltrami operator (LBO)~\cite{do1976differential}. Particularly, we facilitate an accelerated attack by operating in a \textit{shared} spectral coordinate system for all shapes in the dataset. The source's distortions are retained by using low-frequency perturbations and additional mesh-related regularizations.

The attack is tested on datasets of human faces~\cite{ranjan2018generating} and animals~\cite{zuffi20173dmenagerie}. We evaluate SAGA using geometric and semantic metrics. Geometrically, we measure the similarity between shapes by comparing the mean curvature of matching vertices. Semantically, we use a classifier to predict the labels of the adversarial reconstructions, and a detector network to demonstrate the difficulty of identifying the adversarial shapes. We also conduct a thorough analysis of the attack and a comprehensive ablation study.

To summarize, we are the first to propose a \textit{geometric} adversarial attack on 3D meshes. Our method is based on low-frequency spectral perturbations and regularizations of mesh attributes. Using these, SAGA crafts adversarial examples that change an AE's output into a different geometric shape.

%% file: figures/teaser/teaser_pdf.tex
\begin{figure}[t!]
\begin{center}
\includegraphics[width=0.95\columnwidth]{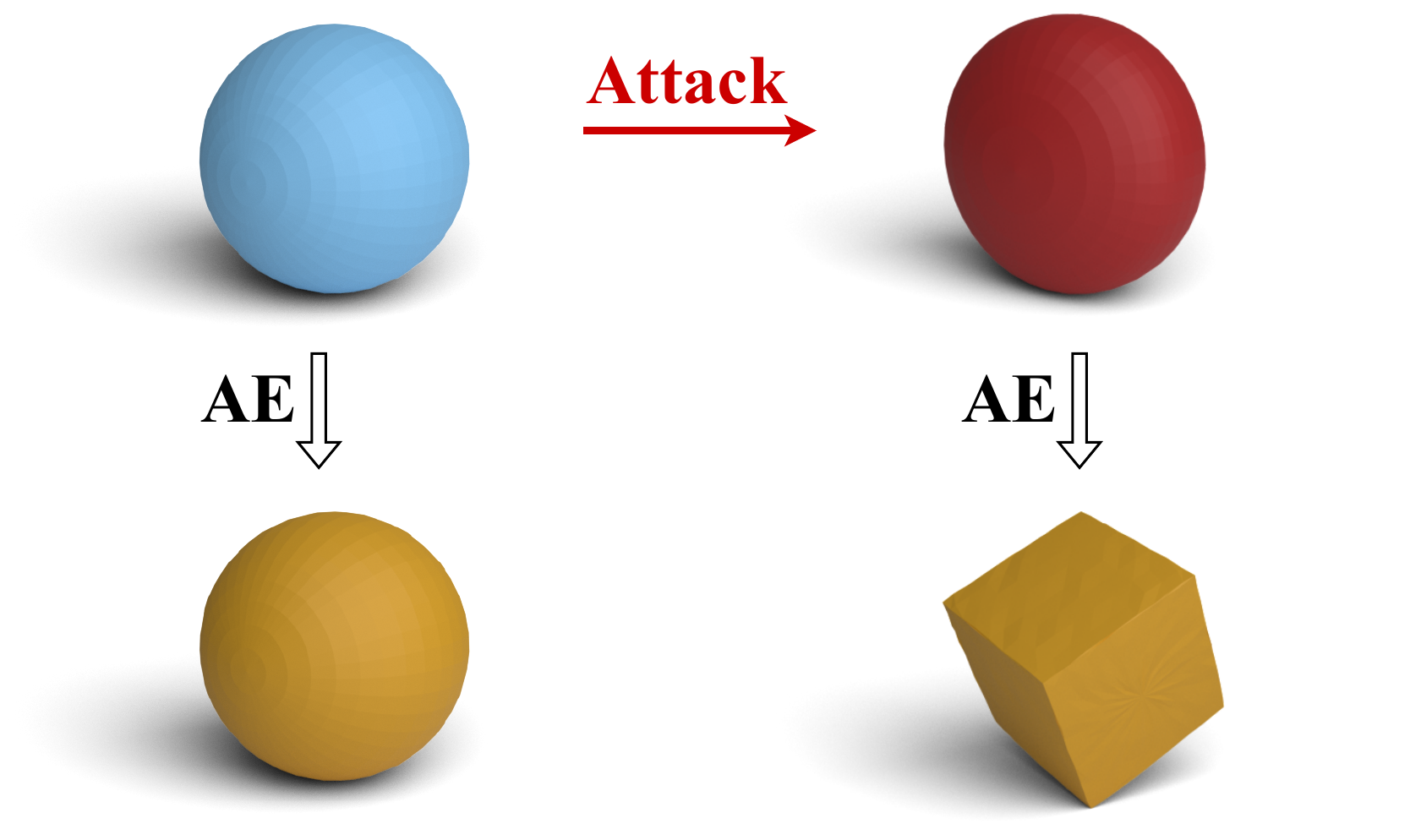}
\vspace{-5pt}
\caption{{\bfseries \newtxt{A result of our geometric mesh attack.}} \newtxt{A mesh of a \textit{sphere} (top left) is perturbed into an adversarial example (top right). While the original mesh is accurately reconstructed by an \textit{autoencoder} (AE) (bottom left), our attack fools the AE and changes the output geometry to a \textit{cube}! (bottom right).}}
\vspace{-12pt}
\label{fig:teaser}
\end{center} 
\end{figure}

%% file: 02_related_work.tex
\section{Related Work} \label{sec:related_work}
%------------------------- Spectral Mesh Analysis
\noindent {\bfseries Spectral mesh analysis.} \quad The vertices and triangular faces of a mesh define a discrete approximation of a 2D surface~\cite{meyer2003discrete}. The spectral analysis of continuous 2D manifolds is derived from the Laplace-Beltrami operator (LBO), which is a generalization of the Laplacian from the Euclidean setting to curved surfaces. The eigenfunctions of the LBO form an orthogonal basis that spans signals upon the shape's surface.

Taubin~\cite{taubin1995asignal} was the first to introduce the spectral analysis of meshes by exploring the notion of a discrete LBO. Pursuing research~\cite{meyer2003discrete, levy2006laplace} suggested using the classic cotangent scheme~\cite{pinkall1993computing} to construct the LBO. In this case, the operator is more robust against differences in mesh discretization. Consequently, the LBO eigenvectors are approximate samples of the continuous eigenfunctions on the vertices of the mesh~\cite{levy2006laplace}. Based on this analysis, we utilized the spectral basis of the mesh to perform our attack.

%------------------------- Mesh Autoencoders
\medskip
\noindent {\bfseries Mesh autoencoders.} \quad Nowadays, a prevailing 3D learning technique employs AE networks that learn to encode geometric shapes into a latent space and reconstruct them. Marin \etal~\cite{marin2020instant} used a multilayer-perceptron (MLP) AE to establish a latent representation of the mesh, and then exploited it in an additional pipeline to recover a shape from its LBO spectrum.

A popular mesh AE was presented by Ranjan \etal~\cite{ranjan2018generating}, where spectral convolution layers and mesh sampling methods achieved promising results on human face data. Further work suggested using spiral convolution operators~\cite{bouritsas2019neural}, while Zhou \etal~\cite{zhou2020fully} used a fully convolutional architecture with a spatially varying kernel to handle irregular sampling density and diverse connectivity. All the mentioned AEs operate on the mesh vertices, assuming a known connectivity, to successfully reconstruct the surface. We used Marin's AE~\cite{marin2020instant} as our victim model, and we explore the attack transferability to the CoMA AE~\cite{ranjan2018generating}.

%------------------------- 3D Adversarial Attacks
\medskip
\noindent {\bfseries 3D adversarial attacks.} \quad In recent years, the research of adversarial attacks on 3D data has expanded, focusing almost entirely on semantic attacks that aim to malfunction classifiers. The literature on semantic adversarial attacks of point clouds is vast. A common approach~\cite{xiang2019generating, hamdi2020advpc} is to refer to the perturbation as shifts or additions of outlier points in the 3D Euclidean space.

On the contrary, semantic mesh attacks often leverage properties derived from the connectivity of the vertices. Belder \etal~\cite{belder2022random} introduced the concept of random walks on the mesh surface to create adversarial examples. Other papers~\cite{mariani2020generating, rampini2021universal} addressed semantic attacks in the spectral domain. Mariani \etal~\cite{mariani2020generating} used band-limited perturbations and extrinsic restrictions to cause misclassifications. Rampini \etal~\cite{rampini2021universal} suggested a universal attack by applying a purely intrinsic regularization on the spectrum of the adversarial shape.

The work most similar to ours is the geometric point cloud attack proposed by Lang \etal~\cite{lang2021geometric}. To our knowledge, this is the only geometric attack on 3D shapes. Lang \etal demonstrated the ability to reproduce a different geometry by feeding an AE with a malicious input shape. However, that work focused on point clouds. It used vertex displacements in the 3D Euclidean space and exploited the lack of connectivity and order to construct adversarial examples.

In contrast, our work is oriented to 3D meshes. Unlike point clouds, meshes have topological constraints. Hence, swaps of vertices' locations or local shifts of vertices are highly noticeable. We leverage the connectivity to operate in the spectral domain where we control global attributes across the shape and better preserve the geometry of the original surface.

%% file: 03_method.tex
\section{Method}  \label{sec:method}

We attack an autoencoder (AE) trained on a collection of shapes from several semantic classes. In each attack, we use a single source-target pair, where the source and target shapes are selected from different classes. Our goal is to find a perturbed version of the source, with minimal distortion, that misleads the AE to reconstruct the target. Ideally, the source's perturbations should be invisible while still altering the AE's output to the geometry of the target shape.

Given an attack setup of a source shape and a target class, we choose, as a pre-processing step, the nearest neighbor shape from the target class in the sense of a Euclidean norm of the difference between matching vertices. Since the AE is sensitive to the geometry of its input, selecting a target that is geometrically similar to the source benefits the attack and reduces the potential magnitude of the perturbation.

In the upcoming subsections, we present a preliminary spectral analysis followed by a description of the spectral domain in which the attack is performed. Then, we define the problem statement and elaborate on the perturbation parameters, the loss function, and the evaluation metrics.

\subsection{Preliminaries}
\noindent {\bfseries Manifolds.} \quad A geometric shape can be described as a 2D Riemannian manifold $\mathcal{X}$ embedded in the 3D Euclidean space $\mathbb{R}^3$~\cite{meyer2003discrete}. Let $\Delta_{\mathcal{X}}$ be the Laplace-Beltrami operator (LBO) of the manifold $\mathcal{X}$, which is a generalization of the Laplacian operator to the curved surface. The LBO admits an eigendecomposition of the shape into a set of discrete eigenvalues \{$\lambda_{i}$\}, known as the spectrum of the shape, and a set of eigenfunctions \{$\phi_{i}$\}, as follows:

\begin{equation} \label{eq:1}
\Delta_{\mathcal{X}}\phi_{i} = \lambda_{i}\phi_{i}.
\end{equation}

The eigenfunctions $\{\phi_{i}\}:\mathcal{X}\xrightarrow{}\mathbb{R}$ form an orthogonal spectral basis of scalar functions. Thus, the Euclidean embedding values of the manifold in the $x,y,z$ axes can be represented as three linear combinations of the spectral basis using a set of corresponding \textit{spectral coefficients} $\{\alpha_{i,x}\}, \{\alpha_{i,y}\}, \{\alpha_{i,z}\}$.

%--------------------------------------- Mesh Graphs
\medskip
\noindent {\bfseries Mesh graphs.} \quad A continuous manifold of a 3D shape can be discretized into a triangular mesh graph $M=(V,F)$. $V\in\mathbb{R}^{n\times3}$ is the vertices matrix, in which each of the $n$ vertices is assigned a 3D Euclidean coordinate. $F\in\mathbb{R}^{m\times3}$ is the triangular faces matrix consisting of $m$ triplets of vertices. We calculate the discrete LBO using the prevailing classic cotangent scheme~\cite{pinkall1993computing}. In this case, the LBO is an $n\times n$ matrix and the eigenvectors are approximated samples of the continuous eigenfunctions on the vertices of the mesh graph~\cite{levy2006laplace}. Let us arrange the eigenvectors as the columns of $\Phi\in\mathbb{R}^{n\times n}$ and the $n$ spectral coefficients of each Euclidean axis as the columns of $A\in\mathbb{R}^{n\times 3}$. Then, the spectral representation of the mesh vertices is given by:

\begin{equation} \label{eq:2}
V=\Phi A. 
\end{equation}

\subsection{Shared Spectral Representation}
The spectral decomposition of a mesh is computationally demanding, and it is restraining the efficiency of our attack. Thus, we propose a novel approach in which the attack is performed in a shared spectral domain. The idea is to represent all the attacked shapes in a shared coordinate system defined by a single set of spectral eigenvectors. This shared basis accelerates the attack by omitting the heavy calculations of a per-shape spectral decomposition.

\input{figures/diagram/diagram_pdf.tex}

%--------------------------------------- Shared Spectral Basis
\medskip
\noindent {\bfseries Shared spectral basis.} \quad 
The spectral decomposition varies between different shapes since the surface of each shape is a unique manifold and its spectral eigenfunctions are defined over its specific geometric domain. However, the geometric resemblance of the shapes in the dataset can be utilized to construct a shared basis of eigenvectors. The idea of a shared set of eigenvectors assures that, practically, the Euclidean coordinates of the vertices of any shape can be spanned by the shared basis with a negligible error.

The shared basis was built as a linear combination of the bases of multiple shapes, which were sampled from different classes. The coefficients of the linear combination were optimized using gradient descent. The loss function was the sum, across all sampled shapes, of the mean-vertex Euclidean distance between the original coordinates and their representation in the shared spectral domain. More details can be found in the supplementary.

%--------------------------------------- Basis Transformation
\medskip
\noindent {\bfseries Basis transformation.} \quad
We denote the shared basis by $\Phi_{shared}\in\mathbb{R}^{n\times n}$, where its columns are the set of $n$ shared eigenvectors. In the new coordinate system, the vertex matrix $V$ of a mesh $M$ can be replaced by the spectral coefficients matrix $A'\in\mathbb{R}^{n\times 3}$ according to:

\begin{equation} \label{eq:3}
V=\Phi_{shared} A'. 
\end{equation}

\noindent Given $\Phi_{shared}$ and $V$, the spectral coefficients are found using least squares. In the following sections, we refer to $A'$ simply as $A$ for ease of notation and assume it was calculated using $\Phi_{shared}$.

\subsection{Attack}
We pose the attack as an optimization problem in a white-box framework, where the AE is fixed. We denote the source mesh taken from class $\mathcal{S}$ by $M_{\mathcal{S}} = (V_{\mathcal{S}}, F_{\mathcal{S}})$, and the target mesh taken from class $\mathcal{T}$ by $M_{\mathcal{T}} = (V_{\mathcal{T}}, F_{\mathcal{T}})$. The spectral representations of $V_{\mathcal{S}}$ and $V_{\mathcal{T}}$ are given by the spectral coefficients matrices $A_{\mathcal{S}}$ and $A_{\mathcal{T}}$, as defined in Equation~\ref{eq:3}. Let us denote by $k$ the number of frequencies we aim to perturb. We add perturbation parameters from $B\in\mathbb{R}^{k\times 3}$ to obtain the adversarial input $A_{adv}$, according to:

\begin{equation} \label{eq:4}
A_{adv}(i)= 
\begin{cases}
    A_{\mathcal{S}}(i)+B(i),& \text{if } i < k \\
    A_{\mathcal{S}}(i),       & \text{otherwise},
\end{cases}
\end{equation}

\noindent where $A_{\mathcal{S}}(i)=[\alpha_{i,x}, \alpha_{i,y}, \alpha_{i,z}]\in\mathbb{R}^{3}$ and $B(i)=[\beta_{i,x}, \beta_{i,y}, \beta_{i,z}]\in\mathbb{R}^{3}$ are the spectral coefficients of frequency $i$ and their perturbation parameters, respectively. Note that the optimized parameters of the attack are the elements of $B$. The resulting adversarial mesh is $M_{adv} = (V_{adv}, F_{\mathcal{S}})$, where $V_{adv} = \Phi_{shared} A_{adv}$. Also, we propose an attack with a multiplicative perturbation, defined as:  

\begin{equation} \label{eq:5}
A_{adv}(i)= 
\begin{cases}
    A_{\mathcal{S}}(i) (1+B(i)),& \text{if } i < k \\
    A_{\mathcal{S}}(i),       & \text{otherwise}.
\end{cases}
\end{equation}

The advantages of operating in the spectral domain are realized by confining the attack to a limited range of low frequencies. By attacking only the low frequencies, we inherently enforce smooth surface perturbations and reduce sharp local changes of the curvature. Consequently, significantly fewer parameters are used compared to a Euclidean space attack where all vertices are shifted. It also offers the flexibility to control the number of optimized parameters.

%--------------------------------------- Problem Statement
\medskip
\noindent {\bfseries Problem statement.} \quad The problem statement is depicted in Figure~\ref{fig:diagram}. The parameters of the perturbation $B$ are optimized according to the following objective:

\begin{equation} \label{eq:6}
\begin{aligned}
\argmin_{B} \quad & \mathcal{L}_{recon}(\widehat{M}_{adv}, M_{\mathcal{T}})+ \mathcal{L}_{reg}(M_{adv}, M_{\mathcal{S}})\\
\textrm{s.t.} \quad & \widehat{M}_{adv}=f_{AE}(M_{adv}),
\end{aligned}
\end{equation}

\noindent where $f_{AE}$ is the AE model and $\widehat{M}_{adv}$ is the reconstruction of $M_{adv}$ by $f_{AE}$. $\mathcal{L}_{recon}$ and $\mathcal{L}_{reg}$ are the loss terms for the target reconstruction and the perturbation regularization, correspondingly. Both terms are further discussed next.

%--------------------------------------- Reconstruction and Regularization Losses
\medskip
\noindent {\bfseries Reconstruction and regularization losses.} \quad The reconstruction of a target shape is achieved by explicitly minimizing the Euclidean distance between the vertices of the AE's output and the vertices of the clean target mesh. Specifically, $\mathcal{L}_{recon}$ is defined as:

\begin{equation} \label{eq:7}
\mathcal{L}_{recon}=\frac{1}{n}\sum_{i=1}^{n} \left\lVert \widehat{V}_{adv}(i) - V_{\mathcal{T}}(i) \right\rVert^{2}_{2}. 
\end{equation}

\noindent where $\widehat{V}_{adv}(i), V_{\mathcal{T}}(i)\in\mathbb{R}^{3}$ are the 3D coordinates of vertex $i$ in meshes $\widehat{M}_{adv}, M_{\mathcal{T}}$, respectively. The sign $\left\lVert \cdot \right\rVert_{2}$ refers to the $l_{2}$-norm. %$\mathcal{L}_{2}$-norm.

To alleviate the distortion of the source shape, we combine the inherent smoothness provided by the spectral perturbations with the $\mathcal{L}_{reg}$ loss. This loss consists of additional mesh-oriented regularizations that are meant to prevent abnormal geometric distortions.

We consider four kinds of regularization measures in $\mathcal{L}_{reg}$, each with a different weight assigned to it. Inspired by Sorkine~\cite{sorkine2005laplacian}, the first term, denoted by $\mathcal{L}_{lap}$, compares the shapes in a non-weighted-Laplacian representation. In this representation, a vertex $V(i)$ is represented by the difference between $V(i)$ and the average of its neighbors. This loss promotes smooth perturbations since it considers the relative location of a vertex compared to its neighbors. Let $I$ be an identity matrix of size $n \times n$, $J$ be the mesh adjacency matrix, and $D = diag(d_{1},...,d_{n})$ be the degree matrix. Then, the non-weighted Laplacian operator, $L_{non}$, is defined as $L_{non} = I - D^{-1} J$, and the vertices matrix is transformed into $\tilde{V} = L_{non} V$. The loss $\mathcal{L}_{lap}$ is defined as: 

\begin{equation} \label{eq:8}
\mathcal{L}_{lap}=\frac{1}{n}\sum_{i=1}^{n} \left\lVert \tilde{V}_{adv}(i) - \tilde{V}_{S}(i) \right\rVert^{2}_{2}.
\end{equation}

The second regularization term, $\mathcal{L}_{area}$, reduces the Euclidean distance between matching vertices, normalized by the total surface area of all the triangles containing the vertex in the clean source shape. The loss $\mathcal{L}_{area}$ retains changes in heavily sampled regions of high curvature, a vital requirement for geometric details preservation. It is defined as:

\begin{equation} \label{eq:9}
\mathcal{L}_{area}=\frac{1}{n}\sum_{i=1}^{n} \frac{1}{area(i)}\left\lVert V_{adv}(i) - V_{\mathcal{S}}(i) \right\rVert^{2}_{2},
\end{equation}

\noindent where $area(i)$ is a weight defined by the sum of the surface area of all the faces containing vertex $i$ in $M_{\mathcal{S}}$. 

Let us denote by $N(M)\in\mathbb{R}^{m\times 3}$ the normal vectors of all the faces of mesh $M$ and by $E(M)\in\mathbb{R}^{d}$ the length of all the edges of mesh $M$, where $d$ is the number of edges. The third and fourth regularization terms in $\mathcal{L}_{reg}$ are denoted by $\mathcal{L}_{norm}$ and $\mathcal{L}_{edge}$, and are defined as follows:

\begin{equation} \label{eq:10}
\mathcal{L}_{norm}=\frac{1}{m}\sum_{i=1}^{m} \left\lVert N(M_{adv})(i) - N(M_{\mathcal{S}})(i) \right\rVert^{2}_{2},
\end{equation}

\begin{equation} \label{eq:11}
\mathcal{L}_{edge}=\frac{1}{d}\sum_{i=1}^{d} |E(M_{adv})(i) - E(M_{\mathcal{S}})(i)|^{2}.
\end{equation}

\noindent The loss $\mathcal{L}_{norm}$ prevents the formation of sharp curves in the adversarial mesh by limiting the deviation of the surface's normal vectors. It is particularly beneficial when the geometric differences between the source and target shapes are coarse. The loss $\mathcal{L}_{edge}$, on the other hand, alleviates local stretches and volumetric changes by keeping the edges' length from changing. Referring to the problem statement in Equation~\ref{eq:6}, we define $\mathcal{L}_{reg}$ as:

\begin{equation} \label{eq:12}
\mathcal{L}_{reg} = \lambda_{l}\mathcal{L}_{lap} + \lambda_{e}\mathcal{L}_{edge} + \lambda_{a}\mathcal{L}_{area} + \lambda_{n}\mathcal{L}_{norm},\\
\end{equation}

\noindent where $\lambda_{l}$, $\lambda_{e}$, $\lambda_{a}$, and $\lambda_{n}$ are the loss terms' weights.

%Exploiting these geometry related attributes revealed practical vulnerabilities that enable the construction of malicious examples.

%--------------------------------------- Evaluation Metrics
\subsection{Evaluation Metrics \label{sec:evaluation_metrics}}
A geometric attack on a mesh AE copes with a built-in trade-off between the need to confine the deformation of the source shape and the requirement to reconstruct the geometry of the different target shape using the AE. We present geometric and semantic quantitative metrics to evaluate these contradicting necessities. 

To geometrically quantify the difference between shapes, we consider a \textit{curvature distortion} measurement, defined as the absolute difference between the mean curvature of matching vertices in the compared shapes. This metric is typically used in semantic mesh adversarial attacks~\cite{mariani2020generating, rampini2021universal}. \newtxt{We use the per-vertex curvature distortion to present heatmaps on the adversarial examples in our visualizations. A complete evaluation of the curvature distortion caused by our attack is reported in the supplementary.} 

%We avoid using other metrics, such as the Euclidean distance since those are less correlated with the mesh representation. Further discussion appears in the supplementary. 

We introduce a semantic evaluation of the adversarial reconstructions and a semantic interpretation of the extent to which the source shape was corrupted. To identify the AE's output, we use a classifier and report the accuracy of labeling the adversarial reconstructions with the target's label. We consider two settings, a targeted and an untargeted classification. In the targeted case, we check whether $\widehat{M}_{adv}$ is labeled as a shape from the target class $\mathcal{T}$. In the untargeted case, we only check if $\widehat{M}_{adv}$ is \textit{not} labeled as a shape from the source class $\mathcal{S}$, which means the semantic identity of the malicious input was altered by the AE.

\input{figures/quiz/quiz_main_pdf.tex}
\footnotetext[2]{\newtxt{(1) original. (2) adversary. (3) adversary. (4) adversary. (5) original. (6) adversary.}}

\newtxt{To appreciate the challenge of detecting adversarial geometric shapes, take the challenge quiz in Figure~\ref{fig:quiz_main}. Can you detect which shapes are clean and which ones are not? We estimate the noticeability of the perturbation by training a detector network in a binary classification task.} The goal is to determine if a certain shape is an adversarial example or not. The detector's accuracy is used as a metric, where a lower score means a better attack.

A dataset of clean source shapes and their perturbed counterparts was constructed for the detection task, where all shapes were originally selected from the AE's test set. The detector was validated and tested using a leave-one-out method, in which shapes from all classes but one were used as the train set. Shapes from the remaining class were split into validation and test sets. For an unbiased comparison, we repeated the experiment multiple times, and each time a different class was excluded for validation and testing. The reported results are an average of all the experiments. A full description of the architecture and the training process appears in the supplementary.

%% file: figures/diagram/diagram_pdf.tex
\begin{figure}[tb!]
\begin{center}
\includegraphics[width=\columnwidth]{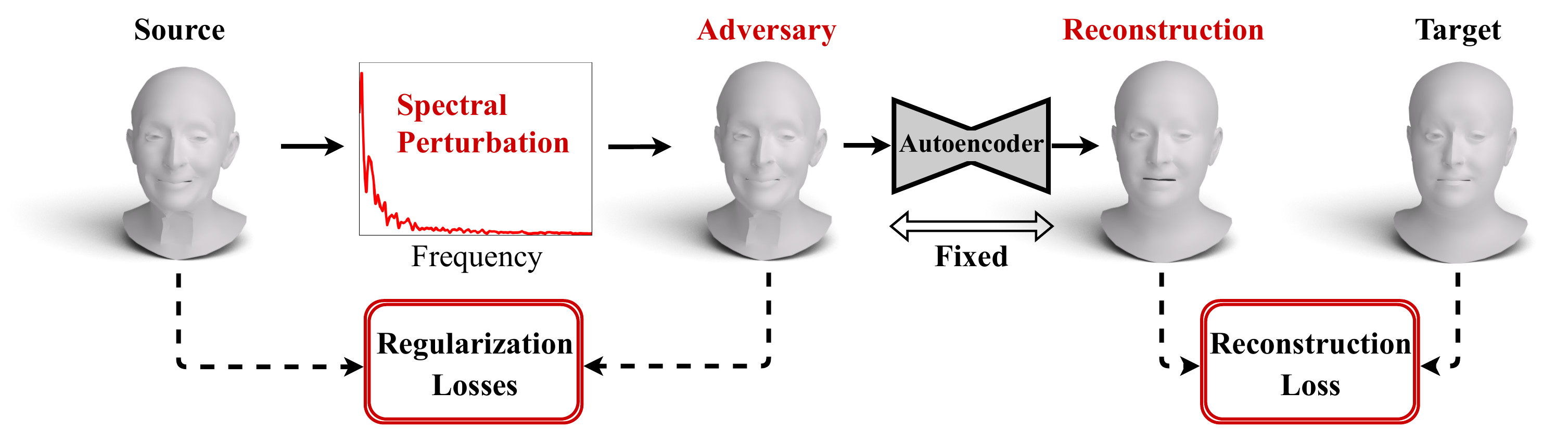}
\caption{{\bfseries The proposed attack framework.} Attack parameters perturb the spectral coefficients of the source shape to craft an adversarial example. The malicious input (Adversary) misleads the AE to reconstruct the geometry of the target mesh. The perturbation is optimized using a loss function that compares the AE's output with the target shape, and regularizes the adversarial shape to preserve the source's geometric properties.}
\vspace{-15pt}
\label{fig:diagram}
\end{center}
\end{figure}

%% file: figures/quiz/quiz_main_pdf.tex
\begin{figure}[tb!]
\begin{center}
\includegraphics[width=0.95\columnwidth]{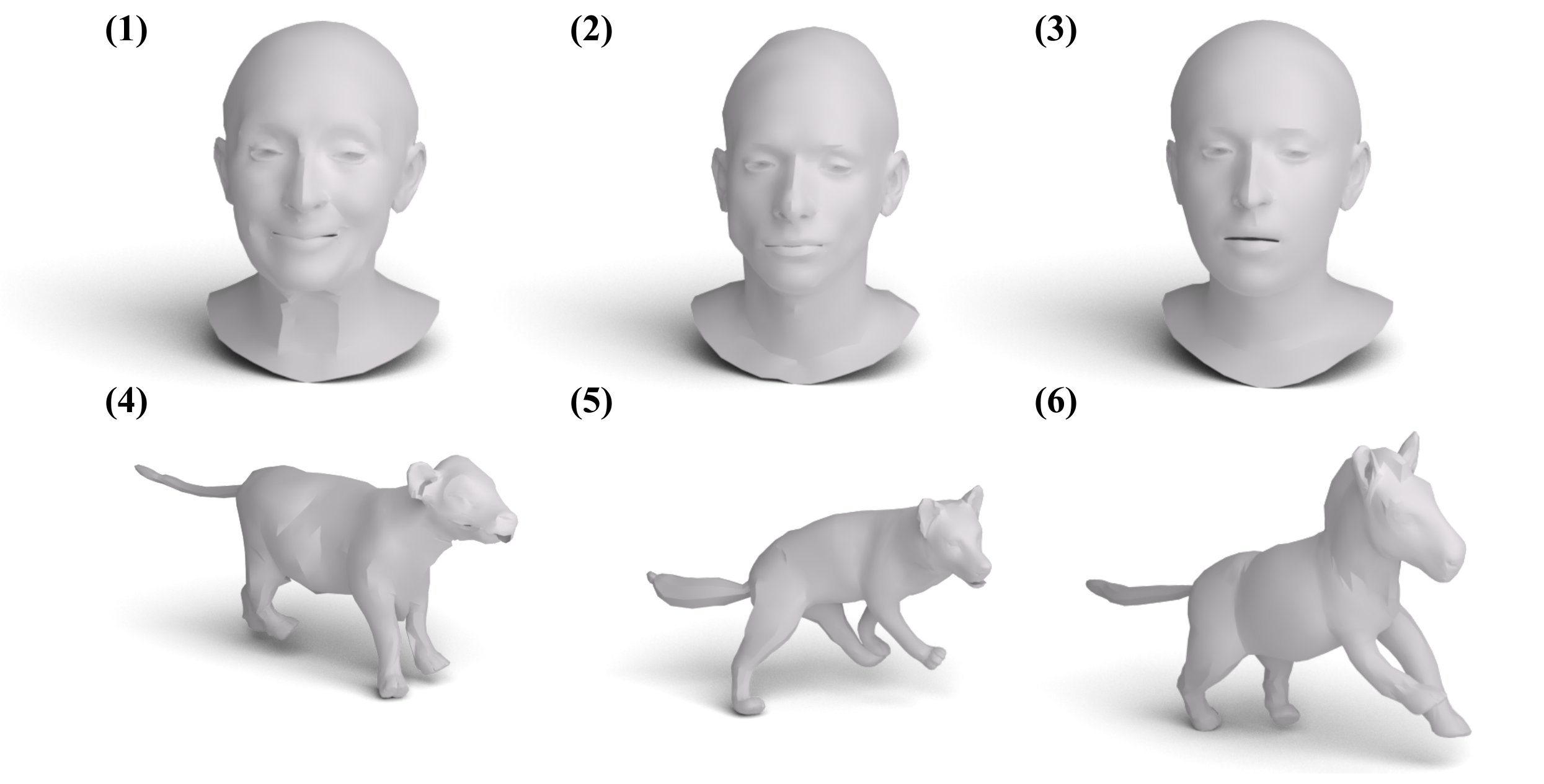}
\caption{{\bfseries \newtxt{Attack detection quiz.}} \newtxt{Which shape is an original mesh from the dataset, and which is an adversarial example of SAGA? The answers can be found in the footnote}\protect\footnote[2]{}.}
\vspace{-12pt}
\label{fig:quiz_main}
\end{center} 
\end{figure}

%% file: 04_results.tex
\section{Results} \label{sec:resutls}

\subsection{Experimental Setup \label{sec:experimental_setup}}
The attack was evaluated on the CoMA dataset of human faces~\cite{ranjan2018generating} and on the SMAL animals dataset~\cite{zuffi20173dmenagerie}. Both datasets are commonly used in the literature~\cite{mariani2020generating, marin2020instant, rampini2021universal, groueix20183dcoded, bouritsas2019neural, belder2022random}. We attacked the mesh AE proposed by Marin \etal~\cite{marin2020instant}. The AE was trained using the same settings as in the original paper for both datasets. During the attack, the AE's weights were frozen, and we used only source and target shapes from the test set.

\medskip
\noindent {\bfseries CoMA.} \quad We used $8325$ examples to train the AE, $926$ for validation, and $1398$ for the test set, where all the sets included instances from $11$ different semantic identities. Shapes from the $12\textsuperscript{th}$ identity were used for an out-of-distribution experiment. During the attack, only the first $500$ frequencies were perturbed with an additive perturbation, as shown in Equation~\ref{eq:4}. The attack parameters were optimized over $500$ gradient steps using Adam optimizer with a learning rate of $0.0001$. We regularized the perturbation using three loss terms, $\mathcal{L}_{lap}$, $\mathcal{L}_{edges}$, and $\mathcal{L}_{area}$, with the corresponding weights $\lambda_{l}=100$, $\lambda_{e}=2$, and $\lambda_{a}=500$.

\medskip
\noindent {\bfseries SMAL.} \quad We used the SMAL parametric model to generate $9918$ shapes of the $5$ animal species, divided into $85\%/10\%/5\%$ for train/validation/test. The variance between classes in the SMAL data required changes in the optimization process compared to the CoMA data. Following Equation~\ref{eq:5}, we performed a multiplicative attack to gain gradual perturbation refinements. We perturbed the eigenvectors of the first $2000$ frequencies. The attack was optimized using the Adam optimizer with a learning rate of $0.01$ over $3000$ gradient steps. We used three regularization terms, $\mathcal{L}_{lap}, \mathcal{L}_{edges}$, and $\mathcal{L}_{norm}$, with the corresponding weights $\lambda_{l}=50, \lambda_{e}=5$, and $\lambda_{n}=0.5$.

\input{figures/attack/attack_pdf.tex}

The attack setup included $50$ source shapes from each class, paired with a single target shape from each of the other classes. This sums up to $50\cdot11\cdot10 = 5500$ attacked pairs for CoMA and $50\cdot5\cdot4 = 1000$ attacked pairs for SMAL. The average attack duration using an Nvidia Geforce GXT 1080Ti was $2.4/13.2$ seconds per pair in the CoMA/SMAL datasets, correspondingly.

We compared our results with the point cloud (PC) attack suggested by Lang \etal~\cite{lang2021geometric}. For a fair comparison, we used the same reconstruction loss as in Equation~\ref{eq:7}. The perturbations were applied as shifts of vertices in the Euclidean space, and we used the Chamfer Distance as the regularization loss, as explained in their paper.

\subsection{\newtxt{Perceptual Evaluation} \label{sec:classifier}}

\input{figures/attack/attacks_comp_pdf.tex}

\newtxt{A visual demonstration of our attack appears in Figure~\ref{fig:attack}. We optimize the changes to the clean source human face such that the AE reconstructs the desired target shape. Restricting the attack to a set of low mesh frequencies, combined with the explicit spatial regularization, maintains the similarity to the source and keeps the natural appearance of the adversarial example.

We compare our attack to the PC geometric attack proposed by Lang \etal~\cite{lang2021geometric}. Figure~\ref{fig:attacks_comp} exhibits a visual comparison. Lang \etal's attack, being adjusted to point clouds, caused a distinctive surface corruption by replacing the order of the vertices. On the contrary, our SAGA reached better target reconstructions with perturbations that preserve the underlying surface.}

We used a PointNet classifier~\cite{qi2017pointnet} to semantically evaluate the adversarial reconstructions. The classifier was trained, validated and tested by the same sets as our victim AE. We trained the model over $1000$ epochs using the same loss function and optimizer as in Rampini \etal's work~\cite{rampini2021universal}.

Table~\ref{tbl:attack_classifier} shows the accuracy obtained from classifying the adversarial reconstructions as the target, in the targeted case, or differently from the source, in the untargeted case. The experiment included all the attacked pairs. We compare our attack with Lang \etal's PC attack~\cite{lang2021geometric} and with the clean target reconstructions.

The results of Table~\ref{tbl:attack_classifier} demonstrate that our attack is also effective on the semantic level. SAGA consistently reached a higher target classification accuracy compared to Lang \etal's attack. On the CoMA dataset, SAGA reached over $99\%$ accuracy in all cases. The results were lower on the SMAL dataset due to the disparity between the different classes. The classifier labeled $67\%$ of SAGA's adversarial reconstructions of animals as the target class. In $82\%$ of the cases, the adversarial reconstructions were classified differently from their source class. In contrast, the PC attack reached a lower accuracy, less than $50\%$ and $75\%$ in the targeted and untargeted settings, respectively.

\input{tables/classifier_evaluation.tex}

\subsection{Attack Detection \label{sec:detector}}

We semantically examined the malicious inputs using a detector network. The detector was separately trained to identify the adversarial shapes of SAGA and the PC attack. We used an MLP architecture to consider the connectivity of the vertices in each mesh. Since the objective of the attack is to create invisible perturbations, a \textit{lower} accuracy rate corresponds to better adversarial examples.

The results of both ours and the PC attack~\cite{lang2021geometric} appear in Table~\ref{tbl:attack_detector}. The detector failed to spot SAGA's perturbations, reaching less than $55\%$ detection accuracy on both datasets. On the other hand, the PC attack was distinctive to the detector. Over $98\%$ of the shapes from CoMA and over $90\%$ of the shapes from SMAL were classified correctly. Therefore, we quantitatively demonstrate the efficiency of SAGA in constructing untraceable malicious inputs. We show that a trained network successfully detects another attack but still fails to identify SAGA's adversarial examples.

\subsection{\tomer{Comparison to Semantic Attacks}} \label{sec:semantic}
The literature on semantic adversarial attacks on 3D meshes is abundant~\cite{xiao2019meshadv, mariani2020generating, rampini2021universal, belder2022random}. Semantic attacks are aimed against classifiers, where adversarial shapes induce misclassifications. An interesting experiment is to check whether a semantic attack is also effective as a geometric attack on an AE. To this end, we applied the semantic \tomer{attacks of Rampini \etal~\cite{rampini2021universal} and Huang \etal~\cite{huang2022shape}} on our data to produce semantic adversarial examples, and \tomer{we} analyzed their impact on the AE.

\input{tables/detector_evaluation.tex}

Using Rampini \etal's framework, we attacked the same animal shapes~\cite{zuffi20173dmenagerie} that were used for SAGA. That is, the attacked set included $250$ animal shapes, consisting of $50$ source shapes from each of the $5$ animal classes. We attacked the pre-trained PointNet classifier~\cite{qi2017pointnet} that was presented in Section~\ref{sec:classifier}. This classifier was also used in Rampini \etal's original paper~\cite{rampini2021universal}, and it was trained, evaluated, and tested using the same sets as our AE. The classifier obtained $99.2\%$ accuracy on the clean shapes. All shapes were originally selected from the classifier's test set.

Although Rampini \etal suggested a universal attack that may be applied to new unseen shapes, we optimized their attack on our specific meshes for a fair comparison. The semantic adversarial meshes were fed through our victim AE and we compare the attack's success rate before and after the AE. The success rate is defined as the accuracy of predicting a different label than the source's label.

A visual demonstration of using \tomer{Rampini \etal~\cite{rampini2021universal}'s} semantic adversarial shapes against the AE is depicted in Figure~\ref{fig:semantic_attack}. The semantic attack altered the labels of its adversarial shapes in $86\%$ of the cases. However, after passing through the AE, the success rate dropped to only $1.6\%$, as opposed to $82.5\%$ of SAGA's reconstructions. Figure~\ref{fig:semantic_attack} demonstrates that the semantic attack fails at the geometric level, as the AE's output remains similar to the source shape. These results show that the semantic attack is ineffective geometrically since it fails to alter the AE's output. In contrast, SAGA is successful in both the geometric and semantic aspects. \tomer{A comparison of our attack to Huang \etal~\cite{huang2022shape}'s semantic attack shows similar results, and it appears in the supplementary.}

\input{figures/semantic_attack/semantic_comp_pdf.tex}

\subsection{\newtxt{Transferability}} \label{sec:transferability}
A common test for an adversarial attack is to check its efficiency on an unseen model. In the following experiment, we \itai{explored} a black-box framework, where the adversarial shapes are used against a different AE than the one they were designed for. We used two unseen AEs. The first has the same architecture as our victim AE but was trained with another random weight initialization. The second is the popular AE proposed by Ranjan \etal~\cite{ranjan2018generating}.

A visual example is presented in Figure~\ref{fig:transferability}. It demonstrates that malicious shapes that were crafted to deceive one AE may change the output of other AEs to the target's geometry. Therefore, SAGA can be transferred to other AEs and still be effective in a black-box setting. More details on the transferred attack can be found in the supplementary.

\input{figures/transferability/transferability_pdf.tex}

\subsection{\tomer{Attacking a Defended AE}} \label{sec:defense_comp}
\tomer{To present} \itai{the} \tomer{robustness of our attack, we tested its efficiency against a defense method. We employed SAGA on an AE defended by the method of Naderi \etal~\cite{naderi2023lpf}. According to their approach, we applied a Gaussian low pass filter on our training set and trained} \itai{the} \tomer{AE with the low pass filtered shapes. Figure~\ref{fig:defense} shows an example of this experiment.}

\tomer{An underlying assumption of the proposed defense~\cite{naderi2023lpf} is that an adversarial attack perturbs the high frequencies of the shape. However, our attack does the exact opposite: it is applied \textit{only} to the low mesh frequencies. Thus, SAGA remains highly effective against the defended AE.}

\input{rebuttal/figures/defense/lpf_defense_pdf.tex}

\subsection{\tomer{Spectral Analysis} \label{sec:spectral_analysis}}
\tomer{We analyze the behavior of the spectral perturbation by measuring its magnitude in each frequency. Recall the notation of the spectral coefficients and their perturbation parameters from Equation~\ref{eq:4}. We define their magnitudes in frequency $i$ as:

\begin{equation} \label{eq:15}
\alpha(i) = \sqrt{\alpha_{i,x}^{2} + \alpha_{i,y}^{2} + \alpha_{i,z}^{2}},
\end{equation}

\begin{equation} \label{eq:16}
\beta(i) = \sqrt{\beta_{i,x}^{2} + \beta_{i,y}^{2} + \beta_{i,z}^{2}}.
\end{equation}

Figure~\ref{fig:coma_combined} shows the average values of $\alpha \in\mathbb{R}^{n}$ and $\beta \in\mathbb{R}^{n}$ over all the attacked pairs, denoted as $\bar{\alpha}$ and $\bar{\beta}$. The perturbation's magnitude follows the natural spectral behavior of the data in both datasets. The graphs demonstrate the attack's emphasis on lower frequencies. By preserving the higher mesh frequencies, SAGA keeps the fine geometric details of the source shape.}

\input{supplementary/figures/analysis/beta_pdf.tex}

\subsection{\newtxt{Additional Experiments}} \label{sec:additional_experiments}
In the supplemental material, we \tomer{analyze the AE's latent space and show the adversarial latent representations}. Also, we conduct an out-of-distribution experiment, where we use a new semantic class that was not part of the AE's training set. We expose the difficulty of reconstructing its unfamiliar figure but the simplicity of altering the geometry of such an unseen identity. As part of a thorough ablation study, we change the regularizations, the number of eigenvectors, and the attacked space. We also present the speed and performance of our attack compared to a spectral attack without a shared basis.

\section{Ethical Considerations} \label{sec:ethical}
Deep Learning for mesh processing has made great progress in recent years. The attack we propose is designed to highlight vulnerabilities in existing methods in hopes of better understanding these models. We acknowledge that such methods can be used negatively in the wrong hands. We hope that shedding light on these vulnerabilities will encourage research on ways to address them.

%% file: figures/attack/attack_pdf.tex
\begin{figure}[tb!]
\begin{center}
\includegraphics[width=\columnwidth]{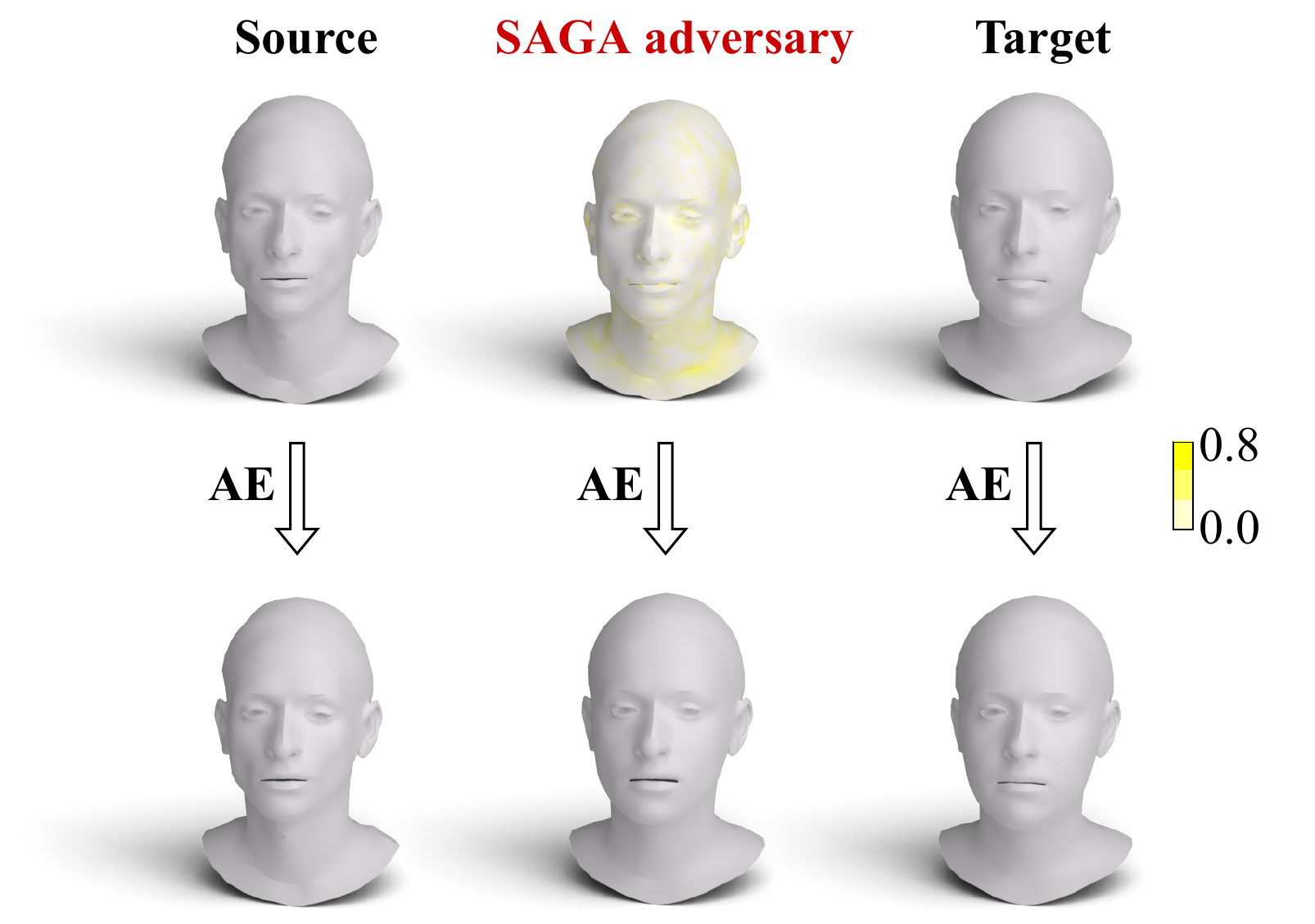}
\caption{{\bfseries Comparison to a clean target reconstruction.} Top row, left to right: the clean source mesh, SAGA’s adversarial example, and the clean target mesh. Bottom row: the reconstructions of the shapes from the top row after passing through the AE. Note that the source has a different identity than the target, with sharper facial features. The heatmap encodes the per-vertex curvature distortion values between the adversarial example and the source mesh, growing from white to yellow. Our mild perturbation of the source human face leads to the reconstruction of a different identity, which is similar to the reconstruction of the clean target.}
\vspace{-10pt}
\label{fig:attack}
\end{center} 
\end{figure}

%% file: figures/attack/attacks_comp_pdf.tex
\begin{figure}[tb!]
\begin{center}
\includegraphics[width=\columnwidth]{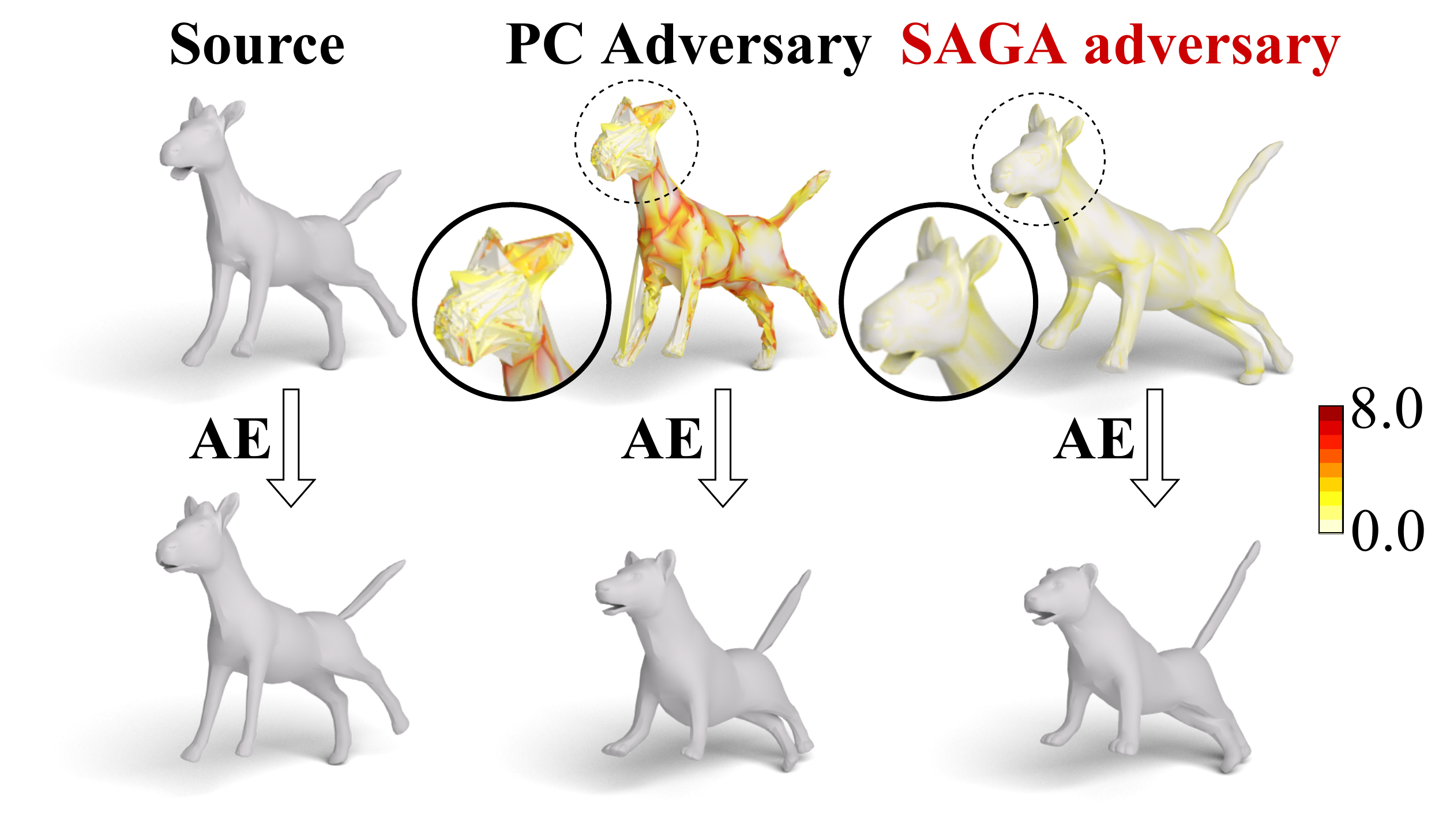}
\caption{{\bfseries Geometric attacks comparison.} Top row, left to right: the clean source mesh (\textit{a horse}), the adversarial example produced by a geometric point cloud (PC) attack~\cite{lang2021geometric}, and SAGA's adversarial example. Bottom row: the reconstructions of the shapes from the top row after passing through the AE. The heatmap encodes the per-vertex curvature distortion values between each adversarial example and the clean source shape, growing from white to red. SAGA's perturbation slightly changes the horse's pose while preserving its geometry. The adversarial horse misleads the AE to reconstruct the geometry of a target \textit{leopard} shape. In contrast, the PC attack causes apparent surface distortions to the source mesh by switching vertices’ locations (as seen in the inset), and its reconstruction lacks the fine-grained details of the target mesh.}
\vspace{-10pt}
\label{fig:attacks_comp}
\end{center} 
\end{figure}

%% file: tables/classifier_evaluation.tex
\begin{table}[tb!]
\centering
\begin{tabular}{@{ } l c c @{ }}
\toprule
Input Type     & Targeted $\uparrow$ & Untargeted $\uparrow$ \\
%\\
\midrule
%\\
Clean target (CoMA)  & 100\% & 100\% \\ 
%\\
\midrule
%\\
PC attack~\cite{lang2021geometric} (CoMA) & 96.22\% & 98.05\% \\
SAGA - ours (CoMA) & {\bfseries 99.31\%} & {\bfseries 99.82\%} \\
%\\
\midrule
%\\
Clean target (SMAL)  & 99.80\% & 100\% \\ 
%\\
\midrule
%\\
PC attack~\cite{lang2021geometric} (SMAL) & 46.70\% & 74.90\% \\
SAGA - ours (SMAL) & {\bfseries 67.00\%} & {\bfseries 82.50\%} \\
\bottomrule
%\\
\end{tabular}
\vspace{0.2cm}
\caption{{\bfseries Semantic interpretation.} The table shows the classification accuracy of the AE's outputs given different inputs. We report the accuracy of labeling the reconstructions as the target class (targeted case) or as any class besides the source class (untargeted case). The adversarial reconstructions of SAGA are compared to those of the point cloud (PC) attack~\cite{lang2021geometric} and to the reconstructions of the clean targets. SAGA consistently outperforms the PC attack on both datasets. The lower accuracy rates on the SMAL~\cite{zuffi20173dmenagerie} dataset stem from the large geometric differences between the source and target shapes.}
\label{tbl:attack_classifier}
\end{table}

%% file: tables/detector_evaluation.tex
\begin{table}[tb!]
\centering
\begin{tabular}{@{ } l c c @{ }}
\toprule
Attack Type     & Detection Accuracy $\downarrow$ \\
%\\
\midrule
%\\
PC attack~\cite{lang2021geometric} (CoMA) & 98.56\% \\
SAGA - ours (CoMA) & {\bfseries 53.69\%} \\
%\\
\midrule
%\\
PC attack~\cite{lang2021geometric} (SMAL) & 90.90\% \\
SAGA - ours (SMAL) & {\bfseries 49.80\%} \\
\bottomrule
%\\
\end{tabular}
\vspace{0.2cm}
\caption{{\bfseries Attack detection.} We report the accuracy of a detector trained to differentiate between adversarial examples and clean inputs. We compare the detection of SAGA to the point cloud (PC) attack~\cite{lang2021geometric}. Details about the training and test procedures appear in Sections~\ref{sec:evaluation_metrics} and~\ref{sec:detector}. Low detection accuracies correspond with a better, unapparent attack. The results demonstrate the difficulty of distinguishing SAGA’s adversarial shapes, in contrast to the distinct recognition of the PC attack.}
\label{tbl:attack_detector}
\end{table}

%% file: figures/semantic_attack/semantic_comp_pdf.tex
\begin{figure}[tb!]
\begin{center}
\includegraphics[width=\columnwidth]{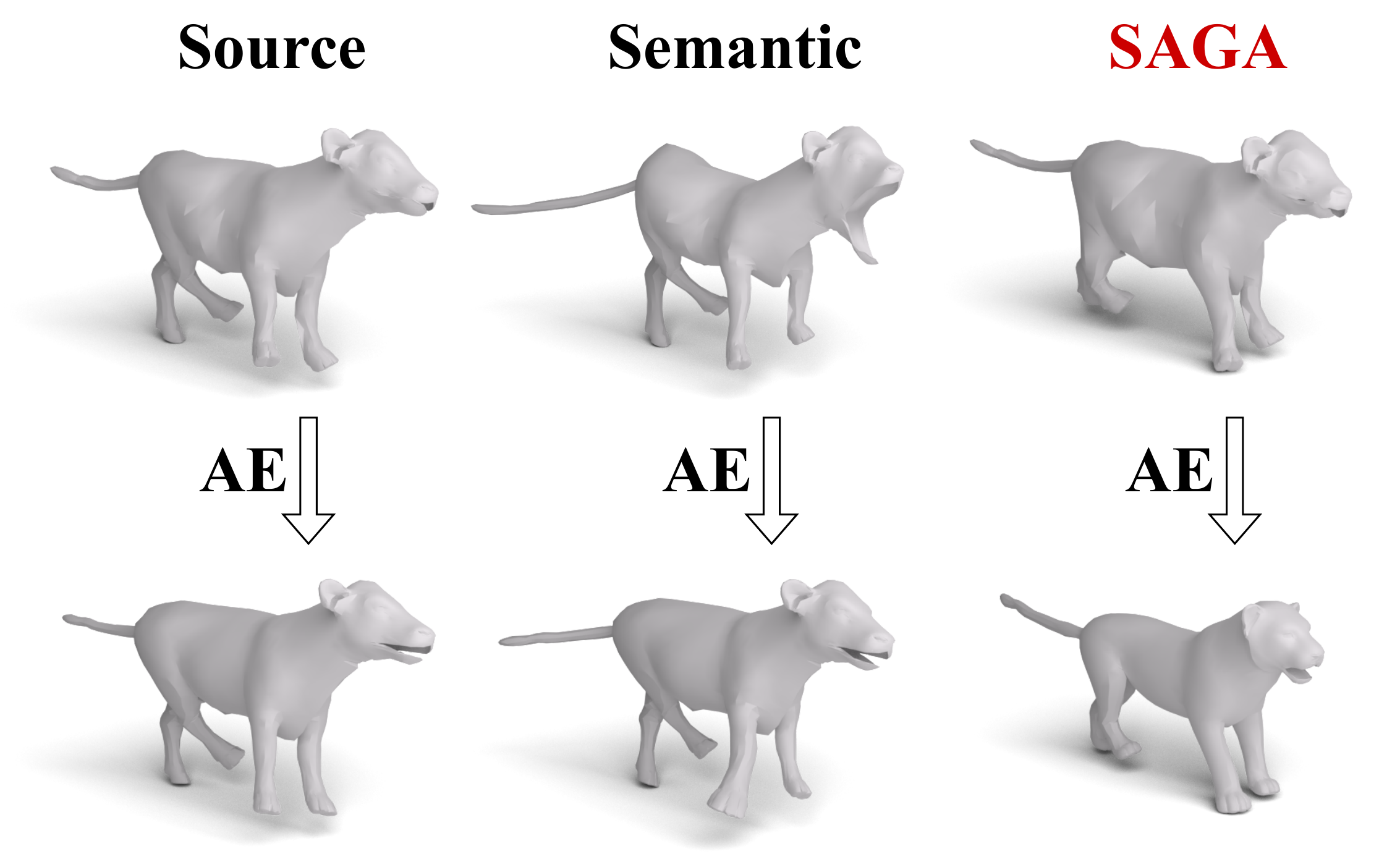}
\caption{{\bfseries \newtxt{A comparison to a semantic attack.}} \newtxt{Top row, left to right: the clean source mesh (\textit{a cow}), the adversarial example produced by a semantic mesh attack~\cite{rampini2021universal}, and SAGA's adversarial example. Bottom row: the reconstructions of the shapes from the top row after passing through the AE. SAGA's adversarial cow successfully misleads the AE to reconstruct the geometry of a target \textit{leopard shape}. However, in contrast to our attack, the reconstructed \textit{shape} of the semantic adversarial mesh remains similar to the source.}}
\vspace{-10pt}
\label{fig:semantic_attack}
\end{center} 
\end{figure}

%% file: figures/transferability/transferability_pdf.tex
\begin{figure}[tb!]
\begin{center}
\includegraphics[width=\columnwidth]{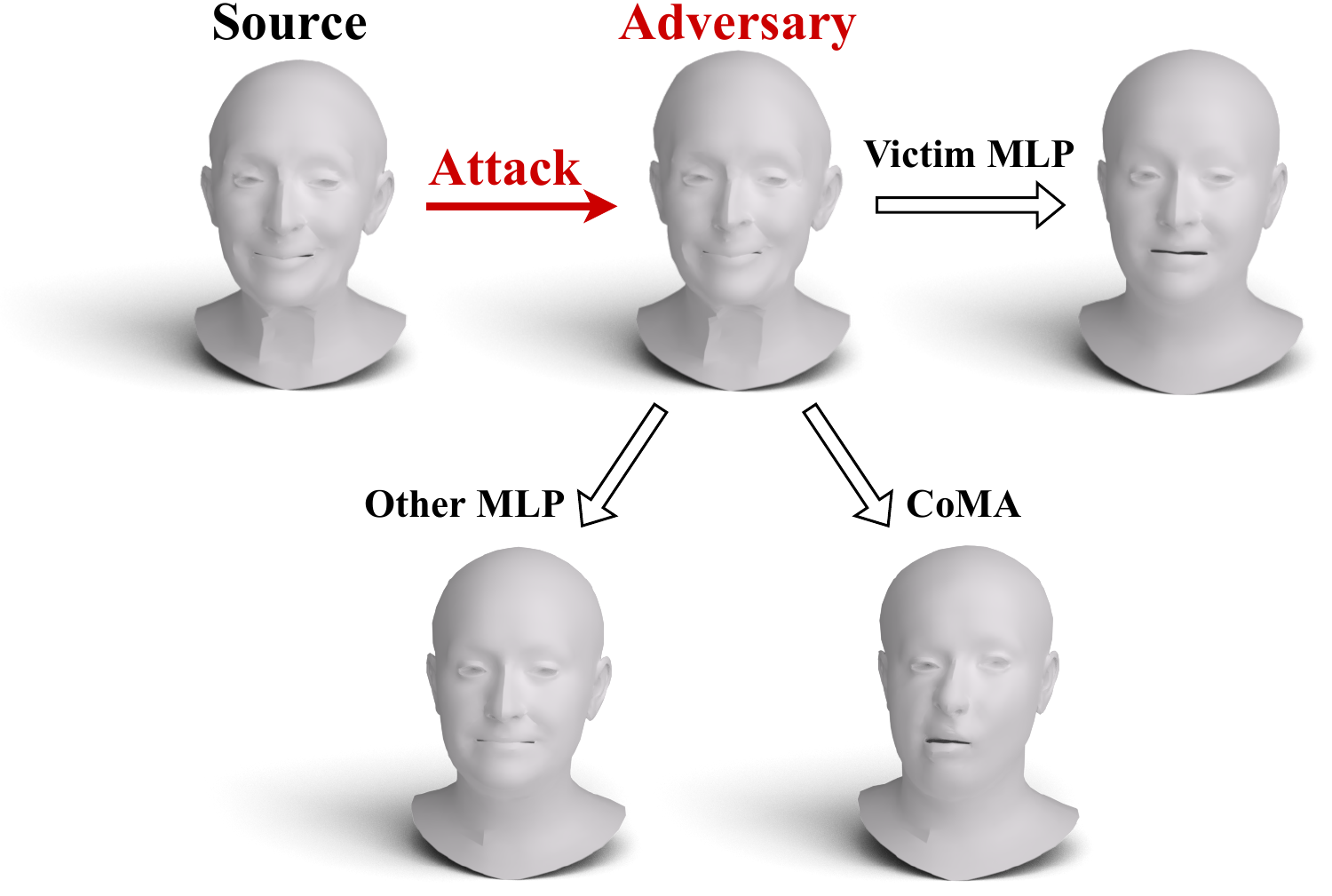}
\caption{{\bfseries Attack transferability.} A source shape (top left) is perturbed by SAGA into an adversarial example (top middle). The adversarial shape passes through three different AEs. The first (top right) is the victim AE used in the attack, with a multilayer perceptron (MLP) architecture (denoted as Victim MLP). The second AE (bottom left) has the same MLP architecture but was trained with a different random weight initialization (denoted as Other MLP). The third (bottom right) is a convolutional AE~\cite{ranjan2018generating} (denoted as CoMA). The three AEs reconstruct the same target identity, and CoMA changes the facial expression of the shape.}
\label{fig:transferability}
\end{center} 
\end{figure}

%% file: rebuttal/figures/defense/lpf_defense_pdf.tex
\begin{figure}[t!]
\centering
\includegraphics[width=\linewidth]{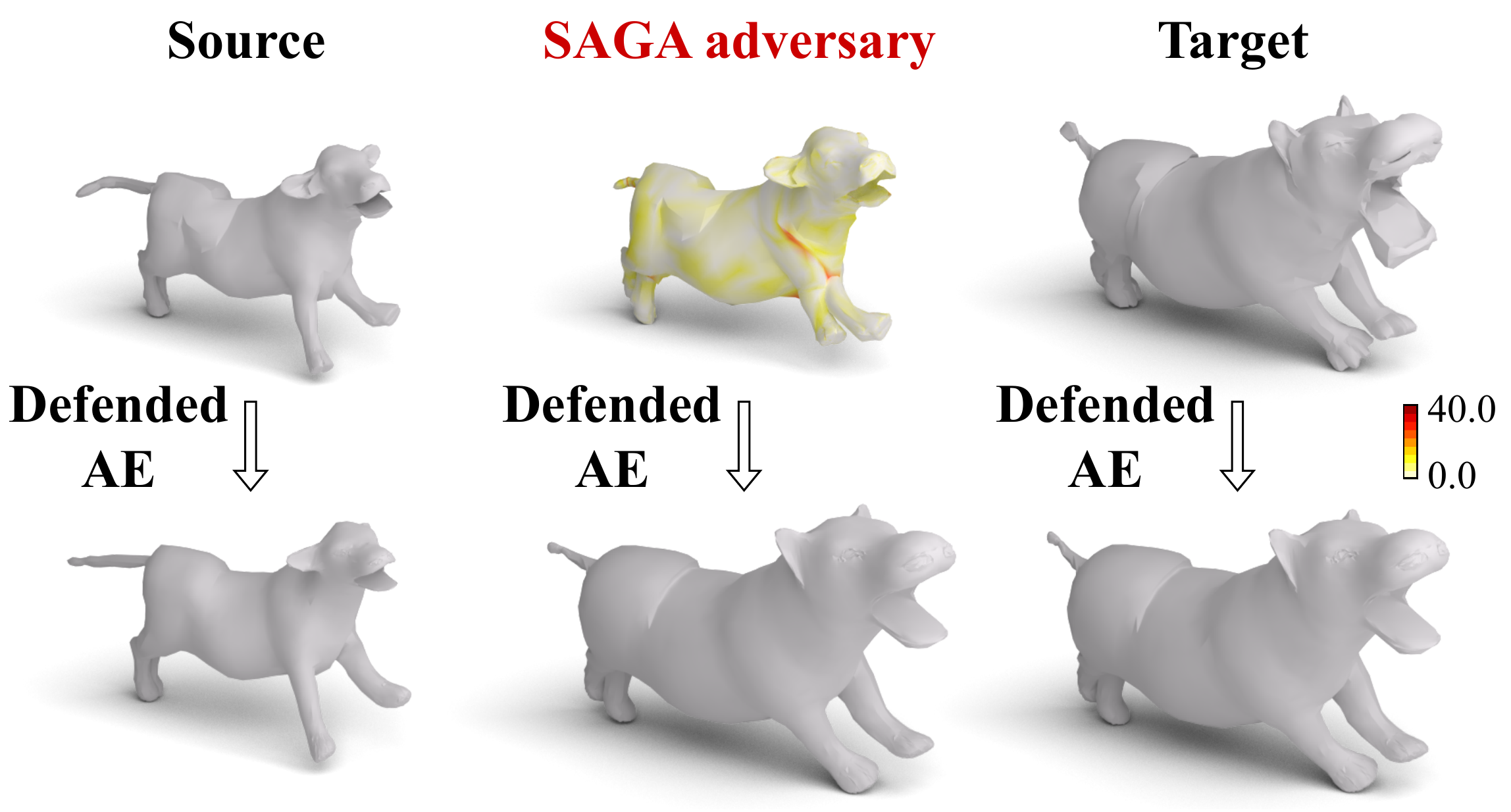}
\caption{\textbf{\tomer{Attack against a defense.}} \tomer{The defended AE was trained on shapes with low frequencies~\cite{naderi2023lpf} and outputs a smoother version of clean inputs (left and right). Our attack is resilient to this defense and successfully alters the reconstructed geometry (center).}}
\label{fig:defense}
\end{figure}

%% file: supplementary/figures/analysis/beta_pdf.tex
\begin{figure}[tb!]
\begin{center}
\includegraphics[width=0.75\columnwidth]{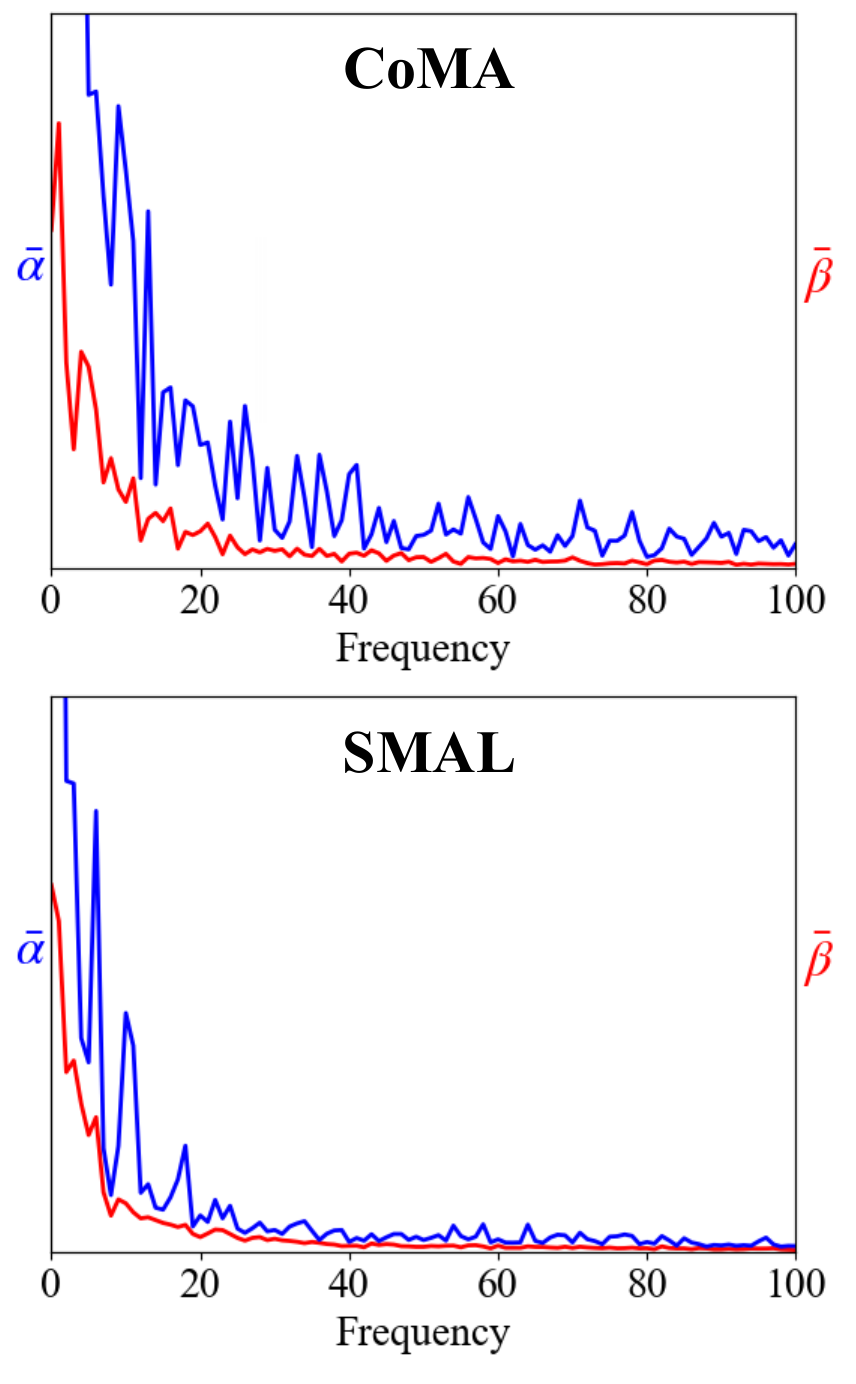}
% \vspace{-5pt}  % to narrow the gap between the fig and caption. you may increase/decrease the -10pt, but make sure that is does not look like you play with the paper template too much...
\caption{{\bfseries \tomer{Spectral analysis}.} \tomer{The graphs show the magnitudes of the spectral coefficients ($\bar{\alpha}$) and their perturbation factors ($\bar{\beta}$) for each frequency, as defined in Equations~\ref{eq:15} and~\ref{eq:16}. The upper graph relates to the results on the CoMA~\cite{ranjan2018generating} dataset, and the lower graph relates to the results on the SMAL~\cite{zuffi20173dmenagerie} dataset. The values are averaged over all the attacked shapes from each dataset. For visual purposes, we truncate the graphs at the frequency 100. The perturbation's magnitude follows the natural spectral behavior of the data. SAGA preserves the higher mesh frequencies, keeping its fine geometric details almost intact.}}
%\vspace{-10pt}  % to narrow the gap between the caption and the paper text below
\label{fig:coma_combined}
\end{center} 
\end{figure}

%% file: 05_conclusions.tex
\section{Conclusions} \label{sec:conclusions}
We introduced a novel geometric attack on a 3D mesh autoencoder (AE). While previous research mostly focused on semantic attacks on classifiers, our method produced malicious inputs that aim to modify the geometry of an AE's output. A previous geometric attack on point clouds utilized the lack of connectivity between points to form adversarial examples. In contrast, a mesh attack is constrained to preserve the delicate structure of the surface to avoid noticeable perturbations. Our method yielded smooth low-frequency perturbations, and leveraged different mesh attributes to regularize apparent malformations.  

We showed that our attack is highly effective in a white-box setting by testing it on datasets of human faces and animals. Semantic and geometric evaluation metrics demonstrated that SAGA's perturbations are hard to detect, while effectively changing the geometry of the AE's output. Our attack outperformed the point cloud attack in all the experiments. Further analysis explored our attack in a black-box scenario, where we demonstrated that SAGA's adversarial shapes are effective against other unseen AEs.

%\section{\tomer{Acknowledgment}} \label{sec:acknowledgment}
\medskip
\tomer{\noindent {\bfseries Acknowledgment.} \quad This work was partly funded by ISF grant number 1549/19.}

%% file: supplementary/supplementary.tex
\clearpage
\appendix

\section*{Supplementary Material}

The following sections include additional information about our attack. In Section~\ref{sec:supp_analysis}, we present a further analysis and conduct additional experiments. In Section~\ref{sec:supp_ablation}, we present a comprehensive ablation study. Section~\ref{sec:supp_settings} includes a complete description of the architectures and the experimental settings. Finally, complementary visual results, observations, and failure examples appear in Section~\ref{sec:supp_results}.

\input{supplementary/01_supp_analysis.tex}

\input{supplementary/02_supp_ablation.tex}

\input{supplementary/03_supp_settings.tex}

\input{supplementary/04_supp_results.tex}

%% file: supplementary/01_supp_analysis.tex
\section{\newtxt{Analysis}} \label{sec:supp_analysis}

\subsection{\newtxt{Geometric Performance} \label{sec:geometric_performance}}

 \newtxt{To geometrically quantify the difference between shapes, we check the absolute difference between the mean curvature of matching vertices in the compared shapes.} We report the average vertex curvature distortion of the adversarial shape compared to the source shape, denoted by $\delta_{\mathcal{S}}$, and of the AE's adversarial reconstruction compared to the target shape, denoted by $\delta_{\mathcal{T}}$. Let us denote by $C(X, Y)$ the average vertex curvature distortion of mesh $X$ compared to mesh $Y$. Then:

\begin{equation} \label{eq:13}
\delta_{\mathcal{S}} = C(M_{adv}, M_{\mathcal{S}}),
\end{equation}

\begin{equation} \label{eq:14}
\delta_{\mathcal{T}} = C(\widehat{M}_{adv}, M_{\mathcal{T}}).
\end{equation}

\noindent We denote the average values of $\delta_{\mathcal{S}}$ and $\delta_{\mathcal{T}}$ over multiple shapes as $\bar{\delta}_{\mathcal{S}}$ and $\bar{\delta}_{\mathcal{T}}$, respectively.

\newtxt{The geometric results of our attack are compared to Lang \etal's point cloud (PC) attack~\cite{lang2021geometric} in Table~\ref{tbl:pc_comparison}. The curvature distortion values of SAGA's adversarial shapes are substantially lower compared to the PC attack, while concurrently, the reconstructions are closer to the target shapes.}

\subsection{\newtxt{Latent Space Analysis}} \label{subsec:supp_more_analysis}

We use t-SNE~\cite{van2008visualizing} to embed the autoencoder's (AE's) latent space into a 2D illustration, as depicted in Figure~\ref{fig:coma_tsne}. The latent representations of clean shapes from different semantic classes are distinctly separated. The representations of several adversarial examples are also displayed. The adversarial examples are encoded into their target's typical latent representation, demonstrating that the attack alters the encoder's predictions.

\subsection{Out of Distribution Attacks} \label{sec:supp_ood}
To extend the scrutiny of our attack's capabilities, we test its performance on new geometric shapes. Our victim AE was trained on shapes from $11$ classes of the CoMA dataset~\cite{ranjan2018generating}. All the previously reported attacks used shapes from a test set that included only these $11$ identities. Here we conduct two experiments, where shapes from a $12\textsuperscript{th}$ \textit{unseen identity} are placed as the source or target of the attack.

\input{supplementary/tables/pc_comparison.tex}

\input{supplementary/figures/analysis/coma_tsne_pdf.tex}

\input{supplementary/figures/out_of_distribution/oods_pdf.tex}

The first experiment included 50 source shapes from CoMA's $12\textsuperscript{th}$ identity, each paired with a target shape from each of the other $11$ identities, leading to $50\cdot11=550$ source-target pairs. In the second experiment, we used 50 shapes from each of the $11$ familiar classes as the source shapes. Their targets were selected from the $12\textsuperscript{th}$ class. Thus, the second experiment also included $550$ attacked pairs. As in previous experiments, the targets were chosen according to the source's nearest neighbor in the target class, in the sense of a mean Euclidean distance between a pair of matching vertices.    

Table~\ref{tbl:oods} shows the average curvature distortion of the perturbed source and its reconstruction in both experiments. We also present the results on the regular source-target pairs from the original attacked set, as shown in Table~\ref{tbl:pc_comparison}. The lowest values of $\bar{\delta}_{\mathcal{S}}$ are obtained when the new shape is used as the source of the attack. It demonstrates that an out-of-distribution shape can be perturbed into an effective malicious input. In this case, $\bar{\delta}_{\mathcal{T}}$ is higher compared to an attack on in-distribution data. When the new identity is placed as the target, the results show substantially higher distortion values of $\bar{\delta}_{\mathcal{T}}$. This result is expected since the AE is required to reconstruct the geometry of an unfamiliar identity. Visualizations of the perturbed new identity are presented in Figure~\ref{fig:oods}.

\input{supplementary/tables/oods.tex}

\input{supplementary/figures/transferability/transferability_pdf.tex}

\subsection{Transferability} \label{sec:supp_transferability}

\input{supplementary/tables/transferability_geometric.tex}

Our method is based on a white-box setting where the AE is used for the optimization process. In this experiment, we explore a black-box setting, where the adversarial shapes are used against a different AE than the one they were designed for. SAGA's adversarial examples of human faces~\cite{ranjan2018generating} were transferred to two other unseen AEs, as described in Section~\ref{sec:transferability} in the main paper.

Figure~\ref{fig:transferability} (in the main paper) and Figure~\ref{fig:supp_transferability} show visual examples of the transferred attack, and the geometric results are reported in Table~\ref{tbl:transferability_geometric}. Geometrically, the curvature distortion values are higher when the attack is transferred to other AEs, especially to the CoMA model~\cite{ranjan2018generating} with a different architecture. However, the visual results demonstrate that the attack is still effective in a black box setting. The reconstructions of the unseen MLP model are visually similar to those of the victim AE. The CoMA model may reconstruct shapes with different facial features than the target's. However, its output facial features are substantially different than those of the source.

We also present a semantic evaluation of the transferred attack. We check the accuracy of predicting that the reconstructed shape has the target's label or a different label than the source's label. The semantic evaluation was conducted on all the attacked shapes, as explained in Section~\ref{sec:classifier}. The results are presented in Table~\ref{tbl:transferability_semantic}. Note that the random accuracy of correctly labeling a shape in a certain class is $9.1\%$ since there are $11$ classes.

The adversarial reconstructions of the unseen MLP AE are predicted to have the target's label in $77\%$ of the cases. In $95\%$ of the cases, the reconstructions are not labeled as the source. Most of the reconstructed shapes from the CoMA model were not classified as their target. However, their classification accuracy remains high in the untargeted case, showing that the input's geometry has changed. We conclude that SAGA remains effective in a black-box setting. When the unseen AE had a different architecture, SAGA was efficient as an untargeted geometric attack.

\input{supplementary/tables/transferability_semantic.tex}

\subsection{\tomer{Comparison to Semantic Attacks}} \label{sec:supp_semantic}

\tomer{Following Section~\ref{sec:semantic} in the main paper, we compare SAGA to the semantic attack proposed by Huang \etal~\cite{huang2022shape}. The attack was applied to the SMAL animal classifier on the same test set we used to evaluate SAGA. A visual comparison is presented in Figure~\ref{fig:rebuttal_comparison}.

The clean shapes before the attack were classified correctly in $99.8\%$ of the cases, as shown in Table~\ref{tbl:attack_classifier}. Huang \etal's attack induced incorrect predictions for $58\%$ of the shapes. However, after passing through the AE, $99\%$ of the reconstructed shapes were correctly classified. In contrast, SAGA's success rate for this setting is $82.5\%$, as shown in Table~\ref{tbl:attack_classifier}. This experiment implies that our attack is more effective than Huang \etal's attack on the geometric level.}

\input{rebuttal/figures/comparison/semantic_comp_pdf.tex}

%% file: supplementary/tables/pc_comparison.tex
\begin{table}[tb!]
\centering
\begin{tabular}{@{ } l c c c c @{ }}
\toprule
Attack Type   & $\bar{\delta}_{\mathcal{S}}$ $\downarrow$  & $\bar{\delta}_{\mathcal{T}}$ $\downarrow$ \\
%\\
\midrule
%\\
PC attack~\cite{lang2021geometric} (CoMA) & 40.17 & 10.67 \\
SAGA - ours (CoMA) & {\bfseries 12.74} & {\bfseries 9.59} \\
%\\
\midrule
PC attack~\cite{lang2021geometric} (SMAL) & 19.64 & 10.62 \\
SAGA - ours (SMAL) & {\bfseries 7.78} & {\bfseries 9.27} \\
%\\
\bottomrule
%\\
\end{tabular}
\vspace{0.2cm}
\caption{{\bfseries \newtxt{Geometric evaluation.}} \newtxt{The metrics for comparison are the average curvature distortion of the adversarial shape concerning the source shape ($\bar{\delta}_{\mathcal{S}}$) and the average distortion of the adversarial reconstruction compared to the target ($\bar{\delta}_{\mathcal{T}}$). The values are averaged over all attacked pairs for each dataset. SAGA is compared to Lang \etal’s point cloud attack~\cite{lang2021geometric}. SAGA's adversarial examples are significantly closer to the source shapes, and yet, their reconstructions are more similar to the target.}}
\label{tbl:pc_comparison}
\end{table}

% \begin{table}[tb!]
% \centering
% \begin{tabular}{@{ } l c c c c @{ }}
% \toprule
% Attack Type   & $\bar{\delta}_{\mathcal{S}}$ $\downarrow$  & $\bar{\delta}_{\mathcal{T}}$ $\downarrow$  & $\bar{\delta}_{\mathcal{T}, N}$ $\downarrow$ \\
% %\\
% \midrule
% %\\
% PC attack~\cite{lang2021geometric_adv} (CoMA) & 40.17 & 10.67 & 1.37 \\
% SAGA - ours (CoMA) & {\bfseries 12.74} & {\bfseries 9.59} & {\bfseries 1.23} \\
% %\\
% \midrule
% PC attack~\cite{lang2021geometric_adv} (SMAL) & 19.64 & 10.62 & 2.34 \\
% SAGA - ours (SMAL) & {\bfseries 7.78} & {\bfseries 9.27} & {\bfseries 2.07} \\
% %\\
% \bottomrule
% %\\
% \end{tabular}
% \vspace{0.2cm}
% \caption{{\bfseries Geometric evaluation.} The metrics for comparison include the average curvature distortion of the adversarial shapes ($\bar{\delta}_{\mathcal{S}}$) and of their reconstructions by the AE ($\bar{\delta}_{\mathcal{T}}$). We also consider the normalized reconstruction distortion relative to the inherent error of the AE ($\bar{\delta}_{\mathcal{T},N}$). SAGA is evaluated compared to Lang's point cloud attack~\cite{lang2021geometric_adv}. The results show that using SAGA the adversarial reconstructions are closer to the target geometry while the source perturbations are significantly lower.}
% \label{tbl:pc_comparison}
% \end{table}

%% file: supplementary/figures/analysis/coma_tsne_pdf.tex
\begin{figure}[tb!]
\begin{center}
\includegraphics[width=0.95\columnwidth]{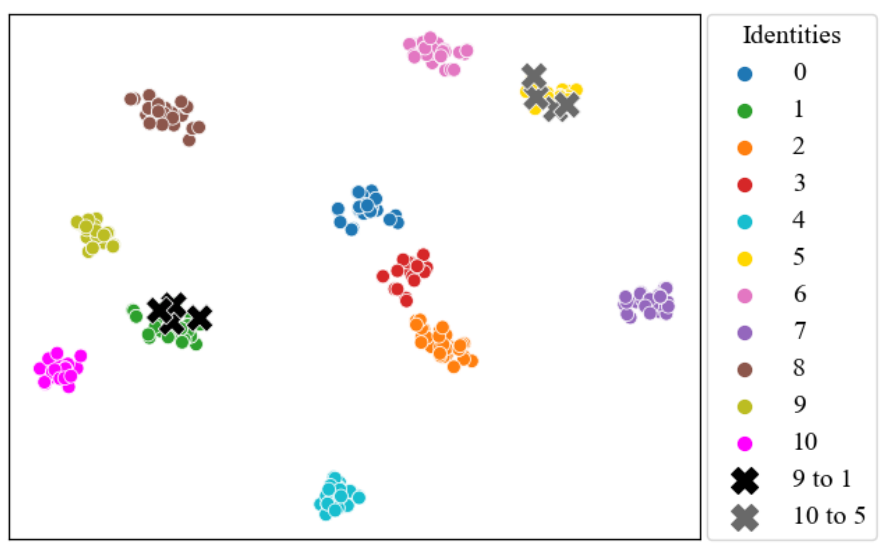}
\caption{{\bfseries A t-SNE~\cite{van2008visualizing} visualization of the latent representation of clean and adversarial shapes.} The figure depicts the latent space of the AE. We present the representations of clean shapes from different semantic classes (identities) in the CoMA~\cite{ranjan2018generating} dataset. Also shown are the representations of adversarial examples. The legend indicates the source and target identities by which the adversarial example was crafted. For instance, `9 to 1' means the source shape had the identity `9' and the target shape was of identity `1'. The adversarial examples are encoded to the typical latent space of their target shape.}
\label{fig:coma_tsne}
\end{center} 
\end{figure}

%% file: supplementary/figures/out_of_distribution/oods_pdf.tex
\begin{figure*}[tb!]
\begin{center}
\includegraphics[width=0.95\linewidth]{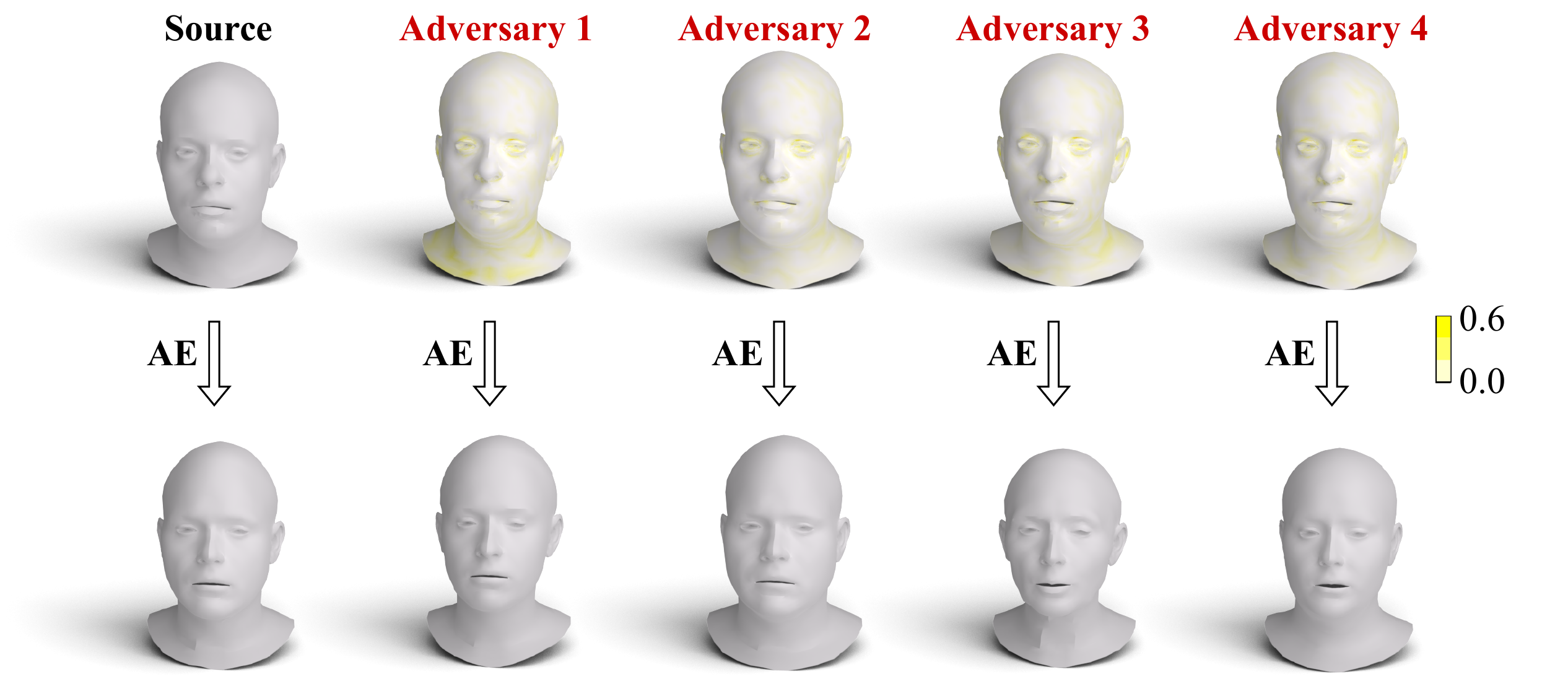}
\caption{{\bfseries An out-of-distribution experiment.} In the top left corner, we show a clean shape of a new identity, unseen by the AE. Other shapes in the top row are four perturbed versions of it, crafted for different targets. The bottom row shows the AE's reconstructions of inputs from the top row. The heatmap encodes the per-vertex curvature distortion values between the adversarial meshes and the source mesh, growing from white to yellow. The figure demonstrates that the AE is vulnerable to shapes of an unfamiliar semantic identity. The adversarial shapes are visually similar to the out-of-distribution source and yet effectively lead to the reconstruction of other target shapes.}
\vspace{-10pt}
\label{fig:oods}
\end{center} 
\end{figure*}

%% file: supplementary/tables/oods.tex
\begin{table}[tb!]
\centering
%\begin{tabular}{@{ } l c c c c @{ }}
\begin{tabular}{l c c c c}
\toprule
Experiment type   & $\bar{\delta}_{\mathcal{S}}$ $\downarrow$  & $\bar{\delta}_{\mathcal{T}}$ $\downarrow$ \\
%\\
\midrule
%\\
Out-of-distribution source & {\bfseries 10.09} & 11.81 \\
Out-of-distribution target & 11.24 & 16.30 \\
In-distribution & 12.74 & {\bfseries 9.59} \\
%\\
\bottomrule
%\\
\end{tabular}
\vspace{0.2cm}
\caption{{\bfseries Out-of-distribution geometric measures.} The results of two experiments are presented. In the first, shapes of a new identity, unseen by the AE, are placed as the source of the attack (Out-of-distribution source). In the second experiment, the shapes are placed as the targets (Out-of-distribution target). We also show the results obtained on familiar source-target pairs (In-distribution), as presented in Table~\ref{tbl:pc_comparison}. The metrics for comparison are the average curvature distortion of the adversarial shape concerning the source shape ($\bar{\delta}_{\mathcal{S}}$) and the average distortion of the adversarial reconstruction compared to the target ($\bar{\delta}_{\mathcal{T}}$). The values are averaged over all attacked pairs. The attack reaches the lowest distortion values when the new identity is placed as the source.}
\vspace{-10pt}
\label{tbl:oods}
\end{table}

%% file: supplementary/figures/transferability/transferability_pdf.tex
\begin{figure}[tb!]
\begin{center}
\includegraphics[width=\columnwidth]{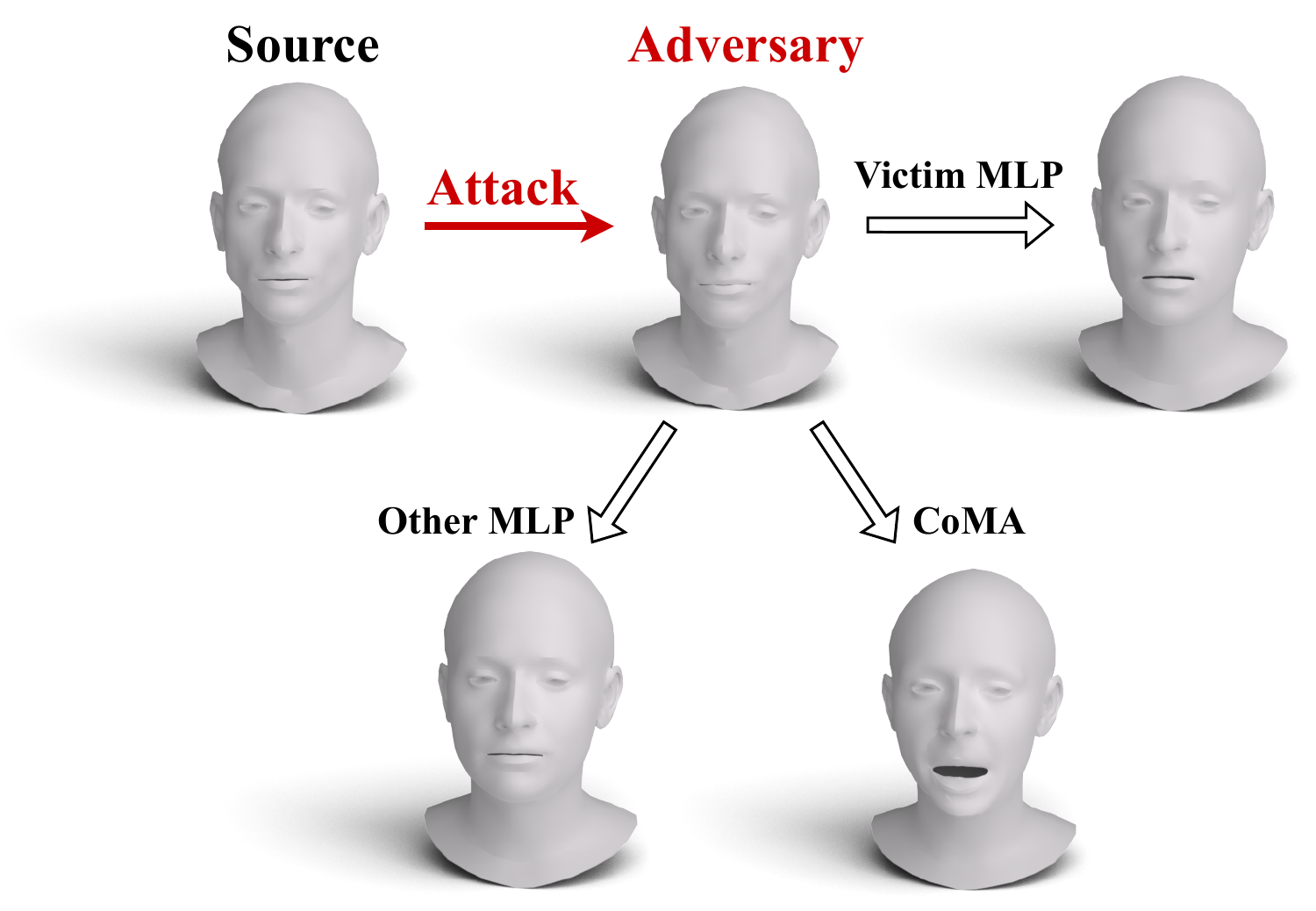}
\caption{{\bfseries Attack transferability.} A source shape (top left) is perturbed by SAGA into an adversarial example (top middle). The adversarial shape passes through three different AEs. The first (top right) is the victim AE of the attack, with an MLP architecture (denoted as Victim MLP). The second AE (bottom left) has the same MLP architecture but was trained with a different random weight initialization (denoted as Other MLP). The third (bottom right) is a convolutional AE~\cite{ranjan2018generating} (denoted as CoMA). Our victim AE and the other MLP AE successfully reconstruct the target geometry. CoMA's reconstruction is different than the target shape. However, its output facial features are different than those of the source.}
\label{fig:supp_transferability}
\end{center} 
\end{figure}

%% file: supplementary/tables/transferability_geometric.tex
\begin{table}[tb!]
\centering
%\begin{tabular}{@{ } l c @{ }}
\begin{tabular}{l c}
\toprule
Autoencoder   & $\bar{\delta}_{\mathcal{T}}$ $\downarrow$ \\
%\\
\midrule
%\\
CoMA~\cite{ranjan2018generating} & 19.44 \\
Other MLP~\cite{marin2020instant} & 12.38 \\
Victim MLP~\cite{marin2020instant} & {\bfseries 9.59} \\
%\\
\bottomrule
%\\
\end{tabular}
\vspace{0.2cm}
\caption{{\bfseries Attack transferability -- geometric evaluation.} SAGA's adversarial shapes were fed through three different AEs. The first is the victim AE with an MLP architecture (Victim MLP). The second AE has the same MLP architecture and a different random weight initialization (Other MLP). The third is a convolutional AE~\cite{ranjan2018generating} (CoMA). We report the average curvature distortion value of the reconstructed mesh ($\bar{\delta}_{\mathcal{T}}$). The experiment was performed on the human face dataset~\cite{ranjan2018generating}, and the values were averaged over all attacked pairs. As expected, we obtained the best results on the victim AE, for which the adversarial shapes were designed. The distortion caused by another AE with the same architecture is higher. The reconstructions of the CoMA model have the highest distortion value.}
%\vspace{-10pt}  % to narrow the gap between the caption and the paper text below
\label{tbl:transferability_geometric}
\end{table}

%% file: supplementary/tables/transferability_semantic.tex
\begin{table}[tb!]
\centering
%\begin{tabular}{@{ } l c c @{ }}
\begin{tabular}{l c c}
\toprule
Autoencoder     & Targeted $\uparrow$ & Untargeted $\uparrow$ \\
%\\
\midrule
%\\
CoMA~\cite{ranjan2018generating} & 33.67\% & 87.02\% \\
Other MLP~\cite{marin2020instant} & 77.04\% & 95.56\% \\
Victim MLP~\cite{marin2020instant} & {\bfseries 99.31\%} & {\bfseries 99.82\%} \\
%\\
\bottomrule
%\\
\end{tabular}
\vspace{0.2cm}
\caption{{\bfseries Attack transferability -- semantic interpretation.} SAGA's adversarial shapes were fed through three different AEs, as described in Table~\ref{tbl:transferability_geometric}. The experiment was performed on the human face dataset~\cite{ranjan2018generating} and included all the attacked meshes. The table shows the classification accuracy of the outputs from each AE. We report the accuracy of labeling the reconstructions as the target class (targeted case) or as any class besides the source class (untargeted case). When transferred to unseen AEs, SAGA remains effective mainly as an untargeted attack.}
\label{tbl:transferability_semantic}
\end{table}

%% file: rebuttal/figures/comparison/semantic_comp_pdf.tex
\begin{figure}[t!]
\centering
\includegraphics[width=\linewidth]{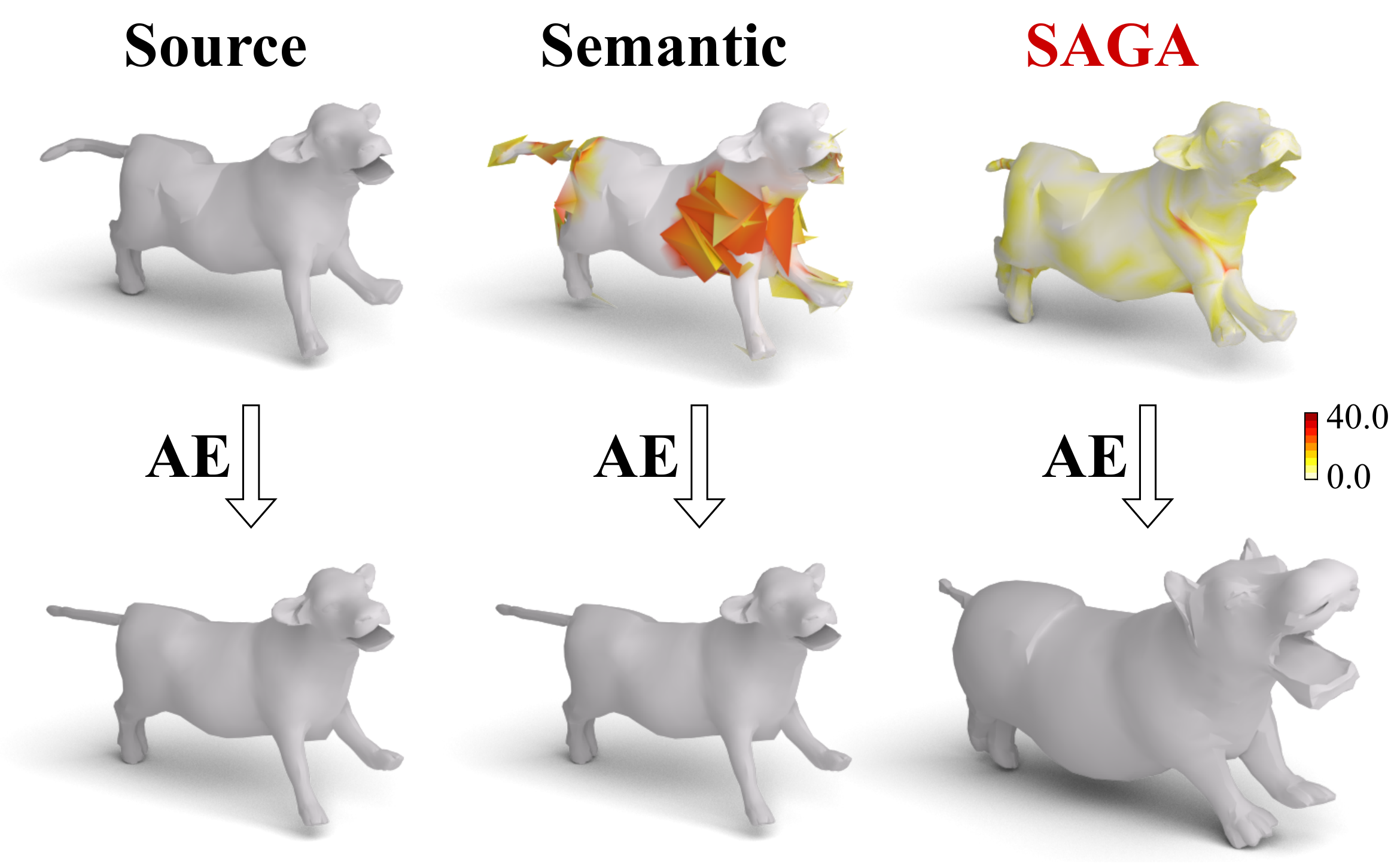}
\caption{\textbf{\tomer{Comparison with an additional semantic attack.}} \tomer{The semantic attack by Huang \etal~\cite{huang2022shape} creates a visible perturbation and is ineffective against the AE. On the contrary, our SAGA better preserves the geometry of the input and effectively changes the reconstructed output shape.}}
\vspace{-5.7pt}
\label{fig:rebuttal_comparison}
\end{figure}

%% file: supplementary/02_supp_ablation.tex
\section{Ablation Study} \label{sec:supp_ablation}

We conduct a thorough ablation study to probe other variants of the proposed framework. Section~\ref{subsec:supp_freq_comp} presents the results of perturbing a different number of eigenvectors and shows the visual effects of band-limited perturbations. In Section~\ref{subsec:supp_reg_comp}, we visually display the contribution of each regularization term. Then, we report the numerical results obtained from various combinations of regularization terms in the loss function. Finally, Section~\ref{subsec:supp_attack_variants} includes a comparison to other attack variants.

\input{supplementary/figures/illustrations/coma_freq_comp_pdf.tex}
 
\subsection{Frequency Study} \label{subsec:supp_freq_comp}

\input{supplementary/figures/analysis/freq_ablation_pdf.tex}

\input{supplementary/figures/illustrations/reg_comp_pdf.tex}

Figure~\ref{fig:coma_freq_comp} displays the visual effects of perturbing different ranges of frequencies. In addition, it shows a Euclidean space attack where the Euclidean coordinates of each vertex are perturbed. The perturbations were produced by running attacks with the reconstruction loss of Equation~\ref{eq:7} and no further regularizations. 

The Euclidean space attack led to a bumpy surface since each vertex can independently be shifted. In contrast, a spectral attack invoked global changes across the surface by perturbing only a fraction of the spanning eigenvectors. By  using a higher range of frequencies, we gradually increase the recurrence of local distortions. Thus, we inherently smooth the perturbations by confining them to a low-frequency range.

In an additional experiment, presented in Figure~\ref{fig:freq_ablation}, we repeat the attack with an increasing number of perturbed eigenvectors. The rest of the attack's parameters are fixed. The average curvature distortions $\bar{\delta}_{\mathcal{S}}$ and $\bar{\delta}_{\mathcal{T}}$ are plotted as a function of the number of used eigenvectors. 

The results reassure the approach of perturbing the low-frequency range. As more degrees of freedom are added to the attack, $\bar{\delta}_{\mathcal{S}}$ increases at a steady paste. However, the improvement in $\bar{\delta}_{\mathcal{T}}$ saturates and even rises in higher frequencies, reflecting their lack of contribution to the attack. In the case of SMAL data, the geometric disparity between classes is large, and the reconstruction of the target geometry is more challenging. Therefore, we used $2000$ eigenvectors to sufficiently assure a low value of $\bar{\delta}_{\mathcal{T}}$. We used $500$ eigenvectors for CoMA data since a further distortion of the source impaired the reconstructions.

\subsection{Regularization Study} \label{subsec:supp_reg_comp}

\input{supplementary/tables/reg_comp.tex}

Figure~\ref{fig:reg_comp} visualizes the contribution of each regularizing term in our loss function. It shows examples of adversarial shapes that were produced by using the reconstruction loss of Equation~\ref{eq:7} and a single regularization term. 

The regularization term $\mathcal{L}_{lap}$ promotes a smooth surface in both datasets. Also, the importance of using $\mathcal{L}_{area}$ to keep subtle geometric details is exemplified, as it preserves curved spots, which are frequently sampled.

The geometric disparity between animal classes is larger than that of human faces. To handle the coarse differences between shapes, we found that $\mathcal{L}_{norm}$ is especially useful to smooth the perturbations. However, its obvious flaw is the inability to prevent local stretches. To compensate, the $\mathcal{L}_{edge}$ loss alleviates stretches and inflations of the shape.

Table~\ref{tbl:reg_comp} contains the quantitative results obtained by using different combinations of regularization terms in the loss function. The inclusion of $\mathcal{L}_{lap}$ (for human faces) and $\mathcal{L}_{norm}$ (for animals) in the loss function leads to the lowest curvature distortion values. Thus, following the visual analysis, we deduce that these are the main smoothing factors for each data type. The other loss terms prevent unnatural stretches and keep the fine geometric curves intact. By regularizing the perturbations with the proposed combination of losses, we keep a natural appearance of the adversarial geometry. 

\subsection{Attack Variants} \label{subsec:supp_attack_variants}

\input{supplementary/tables/euclidean_comp.tex}

We complete our ablation study with a comparison of several variants of the attack. The first is a Euclidean space attack presented in Table~\ref{tbl:euclidean_comp}. In this case, the perturbations occur in Euclidean space, where attack parameters are added to the coordinates of every vertex independently. The loss term and the rest of the attack's settings remain the same. The Euclidean variant shows better geometric measures than SAGA on human faces and mixed results on animals. However, its main drawback is the fixed number of attack parameters. Unlike our spectral method, there is no flexibility to change the number of parameters, and their fixed number is significantly higher.

In a second attack variant, we check the performance of a spectral attack without using a shared coordinates system for all shapes. Without such a prior spectral representation, the Laplace-Beltrami operator (LBO) of the source shape and its eigenvectors are calculated during the attack. Then, the perturbations occur in the spectral basis of each source shape. The loss function and the rest of the attack's settings remain the same. We report the average curvature distortion values in Table~\ref{tbl:self_basis_comp}.

\input{supplementary/tables/self_basis_comp.tex}

As expected, the results are improved when the perturbations occur in the spectral domain of the specific source shape. However, finding the LBO and its eigenvectors is computationally demanding. The time duration of attacking each source-target pair increases by a factor of about $12$ and $3$ on the CoMA and SMAL datasets, respectively.

A third attack variant is presented in Table~\ref{tbl:random_targets}. Here we evaluate SAGA on a new set of attacked pairs, where the targets are selected randomly. Recall that in our main experiment, we pick the target shape according to the source's nearest neighbor in the target class. Here we compare the nearest-neighbor choice with a random selection of a shape from the target class. The overall results show that picking a nearest-neighbor target improves the attack. It reduces the curvature distortion of the adversarial shape and improves its targeted reconstruction. Although the results on human faces~\cite{ranjan2018generating} show a slight improvement in the adversarial reconstructions, we chose the nearest-neighbor setting since the source distortion is lower.

\input{supplementary/tables/random_targets.tex}

%% file: supplementary/figures/illustrations/coma_freq_comp_pdf.tex
\begin{figure}[tb!]
\begin{center}
\includegraphics[width=\columnwidth]{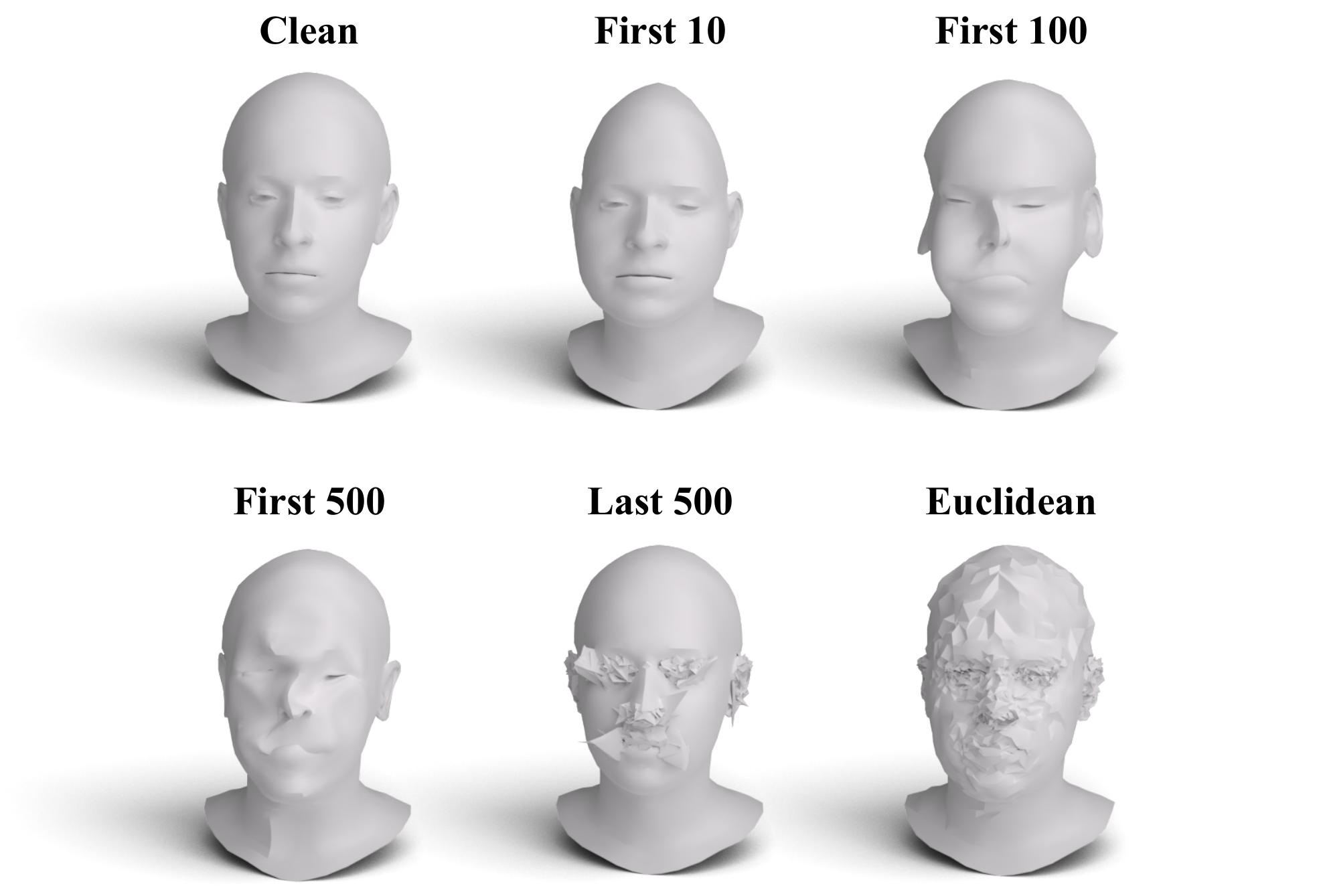}
\caption{{\bfseries Perturbations of different frequencies.} The figure depicts a clean mesh and five perturbed versions of it. The perturbations were produced by running an attack with the reconstruction loss of Equation~\ref{eq:7} and no further regularizations. The term ``First 10/100/500" refers to spectral perturbations of only the first 10/100/500 eigenvectors. The term ``Last 500" means we perturbed only the eigenvectors of the last 500 frequencies in the shared spectral basis. The ``Euclidean" title refers to a Euclidean attack, where we directly perturbed the Euclidean coordinates of each vertex instead of the spectral eigenvectors. Perturbations of higher frequencies gradually increase the recurrence of local distortions. The Euclidean space attack leads to a bumpy surface since each vertex can be shifted independently.}
\label{fig:coma_freq_comp}
\end{center} 
\end{figure}

%% file: supplementary/figures/analysis/freq_ablation_pdf.tex
\begin{figure}[tb!]
\begin{center}
\includegraphics[width=0.95\columnwidth]{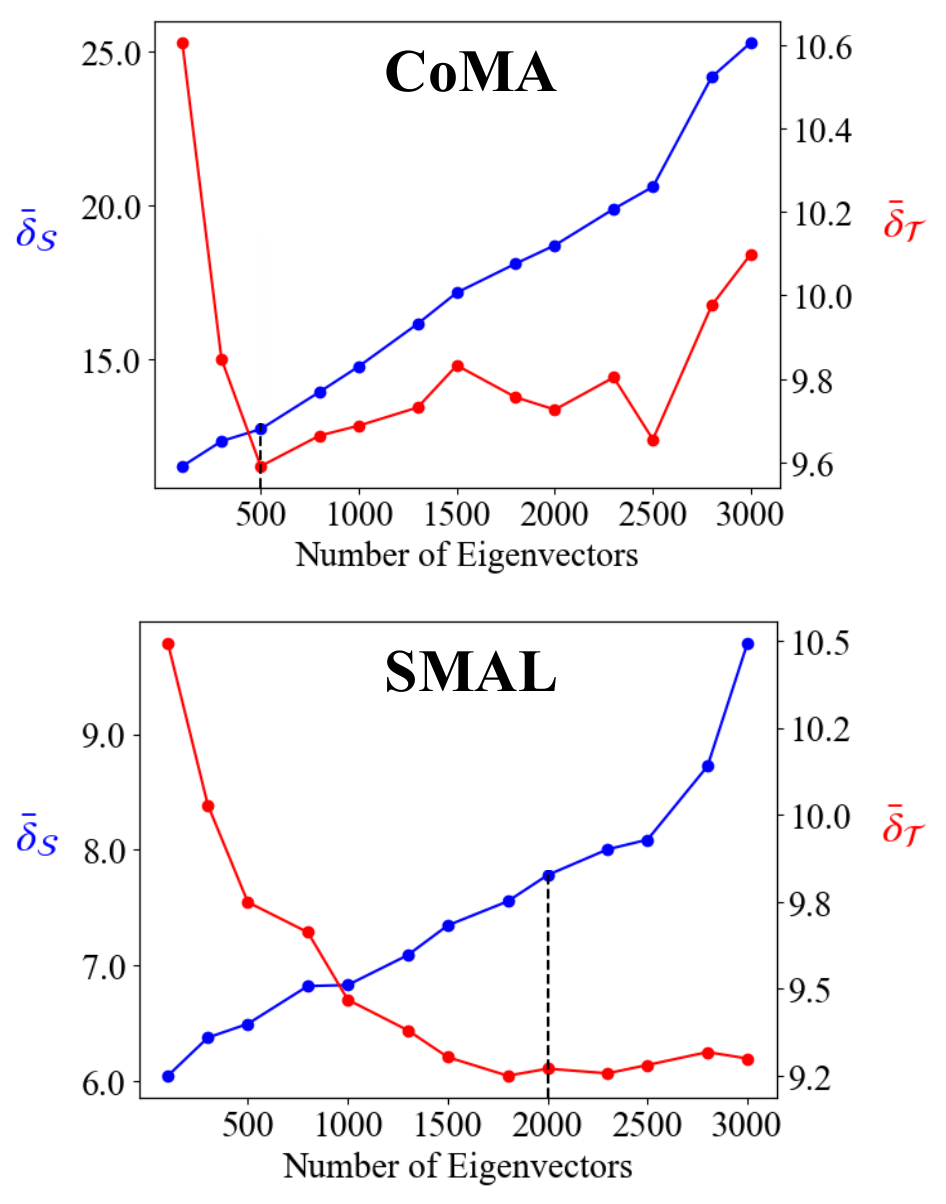}
\caption{{\bfseries Evaluation of a different number of perturbed eigenvectors.} The plots show the average curvature distortions $\bar{\delta}_{\mathcal{S}}$ and $\bar{\delta}_{\mathcal{T}}$ of attacks with a different number of perturbed frequencies. The upper and lower graphs depict the results obtained on the CoMA~\cite{ranjan2018generating} and SMAL~\cite{zuffi20173dmenagerie} datasets, respectively. The horizontal axis denotes the number of consecutive eigenvectors used in the attack, starting from frequency $0$. The dashed vertical line indicates the operating point of our main experiment. By perturbing more eigenvectors, we increase $\bar{\delta}_{\mathcal{S}}$ at a steady paste. However, the improvement in $\bar{\delta}_{\mathcal{T}}$ saturates and even deteriorates in higher frequencies, reflecting their lack of contribution to the attack.} 
\label{fig:freq_ablation}
\end{center} 
\end{figure}

%% file: supplementary/figures/illustrations/reg_comp_pdf.tex
\begin{figure*}[tb!]
\begin{center}
\includegraphics[width=0.95\linewidth]{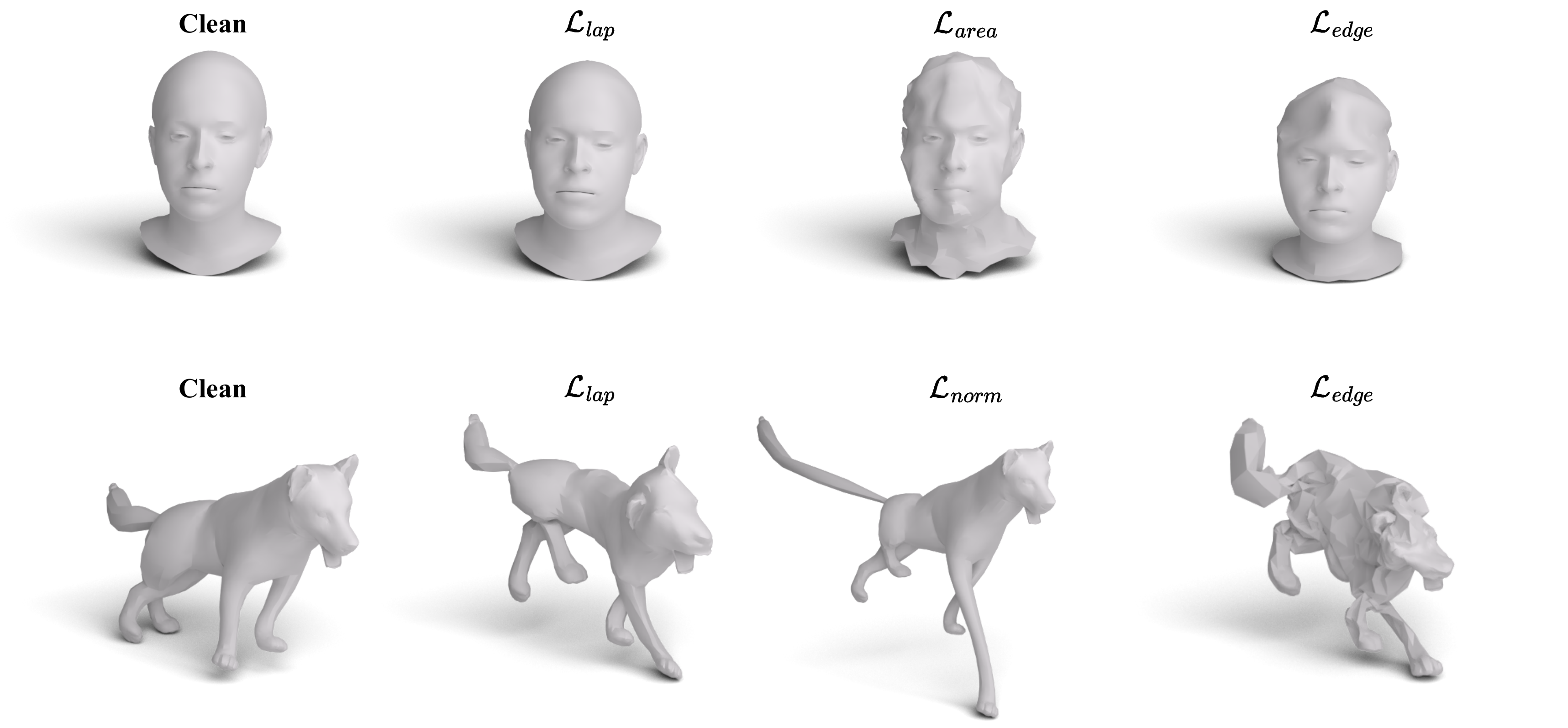}
\caption{{\bfseries Effects of regularization losses.} The figure depicts two clean meshes from the CoMA~\cite{ranjan2018generating} and SMAL~\cite{zuffi20173dmenagerie} datasets (the leftmost column) and perturbed versions of them (the other columns). The perturbations were produced by running an attack with the reconstruction loss of Equation~\ref{eq:7} and different terms for the regularization loss. Each example was regularized using a single loss, referring to the losses defined in Equations~\ref{eq:8},~\ref{eq:9},~\ref{eq:10}, and~\ref{eq:11}. In the CoMA dataset, $\mathcal{L}_{lap}$ promotes a smooth surface. The $\mathcal{L}_{area}$ loss preserves curved areas, which are sampled more frequently. $\mathcal{L}_{norm}$ plays an important role in promoting a smooth surface in the SMAL dataset. However, its obvious flaw is the inability to prevent local stretches. To compensate, $\mathcal{L}_{edge}$ alleviates stretches and inflation of the shape, but it does not retain smoothness by itself. The proposed combination of losses in our SAGA preserves a natural-looking figure of the perturbed shape.}
\label{fig:reg_comp}
\end{center} 
\end{figure*}

%% file: supplementary/tables/reg_comp.tex
\begin{table*}[tb!]
\centering
%\begin{tabular}{@{ } l c c c c c c c c l c c c c @{ }}
\begin{tabular}{c c c c c c c c c c c c c c}
\toprule
%\multicolumn{7}{c}{CoMA~\cite{ranjan2018generating}} & \multicolumn{7}{c}{SMAL~\cite{zuffi20173dmenagerie}} \\
\multicolumn{5}{c}{CoMA~\cite{ranjan2018generating}} &
& & & &
\multicolumn{5}{c}{SMAL~\cite{zuffi20173dmenagerie}} \\
%\\
\midrule
%\\
$\mathcal{L}_{lap}$ & $\mathcal{L}_{edge}$ & $\mathcal{L}_{area}$ & $\bar{\delta}_{\mathcal{S}}$ $\downarrow$ & $\bar{\delta}_{\mathcal{T}}$ $\downarrow$ & & & & &
$\mathcal{L}_{lap}$ & $\mathcal{L}_{edge}$ & $\mathcal{L}_{norm}$ & $\bar{\delta}_{\mathcal{S}}$ $\downarrow$ & $\bar{\delta}_{\mathcal{T}}$ $\downarrow$\\
%\\
\midrule
%\\
\checkmark &  &  & {\bfseries 10.99} & 9.80 &  &  &  &  & \checkmark &  &  & 8.54 & 11.83 \\
 & \checkmark &  & 19.91 & 15.52 &  &  &  &  &  & \checkmark &  & 17.14 & 11.18 \\
 &  & \checkmark & 54.31 & 10.67 &  &  &  &  &  &  & \checkmark & 7.42 & 9.99 \\
%\\
\midrule
%\\
\checkmark & \checkmark &  & 11.09 & {\bfseries 9.24} &  &  &  &  & \checkmark & \checkmark &  & 9.50 & 9.84 \\
 & \checkmark & \checkmark & 21.25 & 14.16 &  &  &  &  &  & \checkmark & \checkmark & 7.95 & 9.46 \\
\checkmark &  & \checkmark & 11.96 & 10.52 &  &  &  &  & \checkmark &  & \checkmark & {\bfseries 7.18} & 10.12 \\
%\\
\midrule
%\\
\checkmark & \checkmark & \checkmark & 12.74 & 9.59 &  &  &  &  & \checkmark & \checkmark & \checkmark & 7.78 & {\bfseries 9.27} \\
%\\
\bottomrule
%\\
\end{tabular}
\vspace{0.2cm}
\caption{{\bfseries Comparison of regularization losses.} The five left columns refer to experiments on human faces~\cite{ranjan2018generating}, and the five right columns refer to experiments on animals~\cite{zuffi20173dmenagerie}. The metrics for comparison are $\bar{\delta}_{\mathcal{S}}$ and $\bar{\delta}_{\mathcal{T}}$, as described in Table~\ref{tbl:oods}. We check different combinations of our regularization components: $\mathcal{L}_{lap}$, $\mathcal{L}_{area}$, $\mathcal{L}_{norm}$, and $\mathcal{L}_{edge}$, described in Equations~\ref{eq:8},~\ref{eq:9},~\ref{eq:10}, and~\ref{eq:11}, respectively. To obtain minimal curvature distortions, the most beneficial regularization terms are $\mathcal{L}_{lap}$ (for human faces) and $\mathcal{L}_{norm}$ (for animals), as their inclusion in the attack's objective leads to the lowest distortion values.}
\label{tbl:reg_comp}
\end{table*}

%% file: supplementary/tables/euclidean_comp.tex
\begin{table}[tb!]
\centering
%\begin{tabular}{@{ } l c c c c @{ }}
\begin{tabular}{l c c c c @{ }}
\toprule
Attack Space   & $\bar{\delta}_{\mathcal{S}}$ $\downarrow$  & $\bar{\delta}_{\mathcal{T}}$ $\downarrow$ & \#Parameters $\downarrow$ \\
%\\
\midrule
%\\
Euclidean (CoMA) & {\bfseries 11.06} & {\bfseries 8.38} & 11793\\
Spectral (CoMA) & 12.74 & 9.59 & {\bfseries 1500}\\
%\\
\midrule
%\\
Euclidean (SMAL) & 9.46 & {\bfseries 8.19} & 11667 \\
Spectral (SMAL) & {\bfseries 7.78} & 9.27 & {\bfseries 6000} \\
%\\
\bottomrule
%\\
\end{tabular}
\vspace{0.2cm}
\caption{{\bfseries Euclidean attack comparison.} SAGA is compared to another attack variant, where the perturbations occur in Euclidean space. All the other attack settings remain the same. The metrics for comparison are $\bar{\delta}_{\mathcal{S}}$ and $\bar{\delta}_{\mathcal{T}}$, as described in Table~\ref{tbl:oods}. The Euclidean variant shows lower distortion values on human faces and mixed results on animals compared to SAGA. However, the number of attack parameters in a Euclidean case is fixed and is significantly larger than in our spectral method.}
\label{tbl:euclidean_comp}
\end{table}

%% file: supplementary/tables/self_basis_comp.tex
\begin{table}[tb!]
\centering
%\begin{tabular}{@{ } l c c c c @{ }}
\begin{tabular}{l c c c c @{ }}
\toprule
Spectral Basis & $\bar{\delta}_{\mathcal{S}}$ $\downarrow$ & $\bar{\delta}_{\mathcal{T}}$ $\downarrow$ & Time (sec) $\downarrow$ \\
%\\
\midrule
%\\
Per-shape (CoMA) & {\bfseries 9.48} & 9.77 & 29.30 \\
Shared (CoMA) & 12.74 & {\bfseries 9.59} & {\bfseries 2.40}\\
%\\
\midrule
Per-shape (SMAL) & {\bfseries 5.33} & {\bfseries 9.08} & 39.08 \\
Shared (SMAL) & 7.78 & 9.27 & {\bfseries 13.20} \\
%\\
\bottomrule
%\\
\end{tabular}
\vspace{0.2cm}
\caption{{\bfseries Spectral bases comparison.} We compare our attack, where perturbations occur in a shared spectral domain, with a similar attack that perturbs each source shape in its self-spectral domain. The rest of the attack's settings remain the same. The metrics for comparison are $\bar{\delta}_{\mathcal{S}}$ and $\bar{\delta}_{\mathcal{T}}$, as described in Table~\ref{tbl:oods}. As expected, we improve the results by representing each source shape in its spectral domain. However, finding a per-shape spectral decomposition is computationally demanding. The duration of attacking each source-target pair is increased by a factor of about 12 and 3 on the CoMA and SMAL datasets, respectively.}
\label{tbl:self_basis_comp}
\end{table}

%% file: supplementary/tables/random_targets.tex
\begin{table}[tb!]
\centering
%\begin{tabular}{@{ } l c c @{ }}
\begin{tabular}{l c c @{ }}
\toprule
Attack Type   & $\bar{\delta}_{\mathcal{S}}$ $\downarrow$  & $\bar{\delta}_{\mathcal{T}}$ $\downarrow$ \\
%\\
\midrule
%\\
Random target (CoMA) & 13.51 & {\bfseries 9.23} \\
NN target (CoMA) & {\bfseries 12.74} & 9.59 \\
%\\
\midrule
Random target (SMAL) & 11.95 & 10.35 \\
NN target (SMAL) & {\bfseries 7.78} & {\bfseries 9.27} \\
%\\
\bottomrule
%\\
\end{tabular}
\vspace{0.2cm}
\caption{{\bfseries Comparison of different target selections.} We evaluate SAGA in two settings of source-target pairs. First, targets are selected randomly from the target class (Random target). In the second alternative setting, the targets are picked according to the source's nearest neighbor in the target class (NN target). The metrics for comparison are $\bar{\delta}_{\mathcal{S}}$ and $\bar{\delta}_{\mathcal{T}}$, as described in Table~\ref{tbl:oods}. The overall results are improved by choosing targets that are geometrically closer to the source shapes.}
\label{tbl:random_targets}
\end{table}

%% file: supplementary/03_supp_settings.tex
\section{Experimental Settings} \label{sec:supp_settings}

\subsection{Autoencoder} \label{subsec:sup_ae}
The victim AE of our attack is the one suggested by Marin \etal~\cite{marin2020instant}. The AE from the original paper included an additional pipeline that maps the Laplace-Beltrami spectrum to the latent representation of the shapes. We omitted this additional pipeline and trained only the spatial model that learns to encode and reconstruct the Euclidean coordinates of the vertices.

Thus, the AE has a multilayer perceptron (MLP) architecture of dimensions: $n\times3\rightarrow300\rightarrow200\rightarrow30\rightarrow200\rightarrow n\times3$, where $n$ is the number of vertices, and the bottleneck size is $30$. All the layers, except the last one, are followed by a $tanh$ activation function. The first component of the loss function is an $\mathcal{L}_{2}$ loss between the reconstructed coordinates and the input. A second component regularizes the weights of every layer, except the last, by summing the $l_{2}$-norm of the weights of each layer and multiplying the sum by a factor of $0.01$. We trained the model using Adam optimizer with a learning rate of $10^{-4}$ and a batch size of $16$ over $2000$ epochs.

\subsection{Classifiers} \label{subsec:sup_classifiers}
The PointNet classifier~\cite{qi2017pointnet} was used to semantically evaluate the reconstructions of the adversarial shapes, and check if they match the targets' labels. Its architecture consists of $4$ point-convolution layers followed by batch normalization and a $ReLU$ activation. The layers' output sizes are $32\rightarrow128\rightarrow256\rightarrow512$. Then, we use a maxpool operation to output a $512$-dimensional vector. The following layers are part of a fully-connected network with output sizes of $512\rightarrow256\rightarrow128\rightarrow64\rightarrow C$, where $C$ is the number of classes. All the fully-connected layers are followed by a $ReLU$ activation function. We trained the classifier with a cross-entropy loss over 1000 epochs. We used a batch size of $6$ and the Adam optimizer with a learning rate of $10^{-3}$.

The detector network was used to distinguish between attacked inputs and clean shapes. Since mesh attacks are obliged to conceal surface distortions, the detector's architecture was designed to consider the order of the vertices. Their order determines the connectivity, which in turn defines the faces of the surface. We used an MLP network with output sizes: $n\times3\rightarrow300\rightarrow200\rightarrow2$. All the layers, except the last one, are followed by a $ReLU$ activation function. The two scalar outputs represent the model's confidence that the input is a malicious shape or a clean one. The decision is made according to the higher value.

The detector was validated and tested using a leave-one-out method, in which shapes from all classes but one were used as the train set. Shapes from the remaining class were split into validation and test sets.

In each experiment of CoMA data, we used $5000$ source-adversarial pairs for training, $250$ pairs for validation, and $250$ pairs for the test set. We reported the average test accuracy of $11$ experiments for SAGA and $11$ experiments for the point cloud (PC) attack~\cite{lang2021geometric}, where each of the $11\textsuperscript{th}$ semantic classes was left for validation and testing. The SMAL data consisted of $800$ source-adversarial pairs for training, $100$ pairs for validation, and $100$ pairs for the test set. The accuracy is the average of $5$ different experiments for SAGA and $5$ others for the PC attack, as the SMAL dataset consists of $5$ semantic classes.

To train the model, we used a cross-entropy loss, a batch size of $6$, and the Adam optimizer with a learning rate of $10^{-5}$. We used $100$ epochs on the data produced by Lang \etal's PC attack. The convergence of the training loss was slower in the case of SAGA's adversarial shapes, where we extended the training to $200$ epochs.

\subsection{Shared Spectral Basis} \label{aubsec:supp_shared_basis}
We built the shared spectral basis using a linear combination of the bases of sampled shapes from each dataset. We sampled $35$ shapes from the CoMA dataset~\cite{ranjan2018generating} and $13$ shapes from the SMAL dataset~\cite{zuffi20173dmenagerie}. The sampled sets included shapes from all classes. Although shapes in the CoMA/SMAL dataset have $3931/3881$ spanning eigenvectors, we found that using $3000$ eigenvectors for a shared basis was sufficiently accurate for both datasets. Hence, we calculated the first $3000$ eigenvectors of each sampled shape and constructed the shared basis using a linear combination:

\begin{equation} \label{eq:17}
\Phi_{shared}=\sum_{i=1}^{P} \gamma_{i} \Phi_{i},
\end{equation}

\noindent where $P$ is the number of sampled shapes, $\Phi_{shared}\in\mathbb{R}^{n\times3000}$ is the shared spectral basis, and $n$ is the number of vertices. $\Phi_{i}\in\mathbb{R}^{n\times3000}$ is the spectral basis of shape $i\in{1,...,P}$, and $\gamma_{i}\in\mathbb{R}$ is its corresponding weight in the linear combination.

The parameters $\gamma_{1},..,\gamma_{P}$ were optimized using gradient descent. We used 50 gradient steps, where the loss function in gradient step $t$, denoted by $L^{t}$, was computed according to:

\begin{equation} \label{eq:18}
L^{t}=\frac{1}{P n}\sum_{i=1}^{P} \sum_{j=1}^{n} \left\lVert V_{i}(j) - (\Phi^{t}_{shared} A^{t}_{i})(j) \right\rVert^{2}_{2}.
\end{equation}

\noindent $V_{i}$ is the vertices matrix of mesh $i$ and $\Phi^{t}_{shared}$ is the shared spectral basis in step $t$, calculated using the weights $\gamma^{t}_{1},...,\gamma^{t}_{P}$. The matrix $A^{t}_{i}\in\mathbb{R}^{3000\times3}$ is the spectral coefficients matrix of shape $i$ in step $t$, calculated using least squares as described in Equation~\ref{eq:3} in the paper. The index $j$ refers to the row of the matrix. To initialize the learned weights, we used the basis of a single shape. For example, in gradient step $t=0$, we set $\gamma^{0}_{1}=1$ and  $\gamma^{0}_{2},...,\gamma^{0}_{P}=0$. We used the Adam optimizer with a learning rate of $10^{-5}$.

We quantitatively evaluate the accuracy of using the shared spectral basis by directly measuring the Euclidean deviation of each vertex in the original mesh compared to its representation by the shared basis. We consider the squared Euclidean norm as the deviation metric and define the overall distance between meshes as the mean vertex deviation. We obtain results on all the meshes from the test set of each dataset and report the average error of a mesh. We compare the representation error of the shared basis with the numerical error of using the spectral basis of each mesh. We also check that these errors are lower than the inherent reconstruction error of the AE. 

This examination shows that the representation error of the shared basis is slightly higher than the numerical error of using the spectral basis of each mesh. In both cases, the average error of a mesh is approximately $9\cdot10^{-9}$ and $8\cdot10^{-7}$ on the CoMA~\cite{ranjan2018generating} and SMAL~\cite{zuffi20173dmenagerie} datasets, respectively. The average reconstruction error of the AE is higher by a factor of $50$ and $10^{3}$ on the corresponding datasets.

\subsection{\newtxt{Geometric Metrics}} \label{subsec:sup_geo_evaluation}

In this work, we quantitatively evaluate the geometric difference between shapes using the curvature distortion metric, as explained in Section~\ref{sec:evaluation_metrics} in the main paper. Another approach is to select the Euclidean distance ($l_{2}$-norm) as the metric to compare meshes. However, we avoided using this metric since it is agnostic to the mesh's topology. A small vertex shift that results in a low Euclidean error, may still cause a noticeable surface distortion, like in cases of interchanging vertices. Moreover, our concealed perturbations often relate to smooth global shifts of vertices across the shape and even pose changes. Such perturbations invoke a high $l_{2}$-norm value while the natural appearance of the shape is preserved.

%% file: supplementary/04_supp_results.tex
\section{Visual Results} \label{sec:supp_results}

\input{supplementary/tables/stability.tex}

\subsection{Adversarial Stability} \label{subsec:stability}

An interesting question about a geometric attack is how stable are its adversarial reconstructions. To answer this question, we check if SAGA's adversarial reconstructions can be re-used by the AE. Particularly, after feeding the AE with an adversarial example, we use the reconstructed output as a new input to the AE. Figures~\ref{fig:coma_stability} and~\ref{fig:smal_stability} show three iterations of the described process on both datasets. A semantic evaluation of the AE's output, after each iteration, is presented in Table~\ref{tbl:stability}. Note that we use the same data and classifier as presented in Section~\ref{sec:classifier}.

The visual results demonstrate that SAGA's adversarial reconstructions are stable, and their geometry is preserved even after encoding and reconstructing them by the AE several times. The classification results in Table~\ref{tbl:stability} statistically back these findings. The reconstructed shapes of human faces~\cite{ranjan2018generating} are classified as their targets in over $99\%$ of the cases. This accuracy remains stable after two more iterations of encoding and decoding. The classification accuracy of the animal shapes~\cite{zuffi20173dmenagerie} deteriorates by $9\%$-$12\%$ after two iterations. However, most of the adversarial reconstructions remain stable.

\subsection{Limitations} \label{sec:supp_failure}

In this section, we elaborate on the limitations of the attack. Naturally, failures tend to appear in cases where the geometries of the source and target shapes are especially distant. In the CoMA dataset~\cite{ranjan2018generating}, the geometric resemblance between shapes is relatively high, and the variety of identities is reflected in delicate changes in the portrait of the face. As a result, the main limitation of our attack is the difficulty to control local deformations of subtle facial features. Figure~\ref{fig:coma_fail} shows a scenario where the adversarial example has unnatural deformations that resulted in a distorted reconstruction.

Alternatively, the geometric differences between classes are substantially larger in the animals' dataset~\cite{zuffi20173dmenagerie}. A typical failure case occurs when the geometry of another animal, which is more similar to the source, is reconstructed instead of the target. Examples of this failure appear in Figure~\ref{fig:smal_fail}. In one case, an adversarial example of a dog led to the reconstruction of a leopard instead of a cow. In another, a perturbed dog resulted in a horse instead of a hippo.

\subsection{Attack Evolution} \label{subsec:attack_evolution}

Figures~\ref{fig:coma_evolution} and~\ref{fig:smal_evolution} present visualizations of the attack's optimization progress in different gradient steps. We present visualizations of human faces~\cite{ranjan2018generating} and animals~\cite{zuffi20173dmenagerie}. The attack converges into a decent solution in the early optimization steps on both datasets. The remaining gradient steps fine-tune the solution gradually. The example in Figure~\ref{fig:smal_evolution} shows that the reconstructed shape gradually turns from a horse into a leopard while changing into a cow in between.

\subsection{More Visual Results} \label{subsec:supp_more visuals}

\input{supplementary/figures/quiz/quiz_answers_pdf.tex}

To further exhibit SAGA, we display more visual results in Figures~\ref{fig:coma_concat_1},~\ref{fig:coma_concat_2},~\ref{fig:smal_concat_1}, and~\ref{fig:smal_concat_2}. \newtxt{In addition, we show the outcome of passing the shapes from the quiz in Figure~\ref{fig:quiz_main} through the AE. That is, we used our AE to encode and reconstruct each of the meshes from the quiz, and the results are presented in Figure~\ref{fig:quiz_recon}. Notice that the clean shapes were accurately reconstructed, and the adversarial shapes effectively misled the AE to reconstruct different geometric shapes.}

\clearpage

\input{supplementary/figures/stability/coma_stability_pdf.tex}

\input{supplementary/figures/stability/smal_stability_pdf.tex}

\input{supplementary/figures/failures/coma_failure_pdf.tex}

\input{supplementary/figures/failures/smal_failure_pdf.tex}

\input{supplementary/figures/evolution/coma_evolution_pdf.tex}

\input{supplementary/figures/evolution/smal_evolution_pdf.tex}

\input{supplementary/figures/attack/coma_concat_1_pdf.tex}

\input{supplementary/figures/attack/coma_concat_2_pdf.tex}

\input{supplementary/figures/attack/smal_concat_1_pdf.tex}

\input{supplementary/figures/attack/smal_concat_2_pdf.tex}

%% file: supplementary/tables/stability.tex
\begin{table}[tb!]
\centering
%\begin{tabular}{@{ } l c @{ }}
\begin{tabular}{l c @{ }}
\toprule
AE Rounds     & Accuracy $\uparrow$ \\
%\\
\midrule
%\\
Iteration 1 (CoMA) & 99.31\% \\
Iteration 2 (CoMA) & 99.49\% \\
Iteration 3 (CoMA) & {\bfseries 99.51}\% \\
%\\
\midrule
%\\
Iteration 1 (SMAL) & {\bfseries 67.00}\% \\
Iteration 2 (SMAL) & 58.10\% \\
Iteration 3 (SMAL) & 55.30\% \\
\bottomrule
%\\
\end{tabular}
\vspace{0.2cm}
\caption{{\bfseries Semantic evaluation of the attack's stability.} We report the classification results of three stability iterations. First, the adversarial shapes are fed through the AE, and we evaluate the reconstructions' labeling accuracy (Iteration 1). Then, the reconstructions are reused as the inputs, and we check the classification of the new outputs (Iteration 2). In the third iteration, the inputs to the AE are the AE's outputs from the second iteration. The table presents the accuracy of labeling the reconstructions as the target class on the CoMA~\cite{ranjan2018generating} and SMAL~\cite{zuffi20173dmenagerie} datasets. The classification accuracy of the adversarial reconstructions remains stable on human faces, even after several rounds of encoding and reconstructing. Although the accuracy on animals decreases after each iteration, most of the adversarial examples remain stable.}
\label{tbl:stability}
\end{table}

%% file: supplementary/figures/quiz/quiz_answers_pdf.tex
\begin{figure}[tb!]
\begin{center}
\includegraphics[width=0.95\columnwidth]{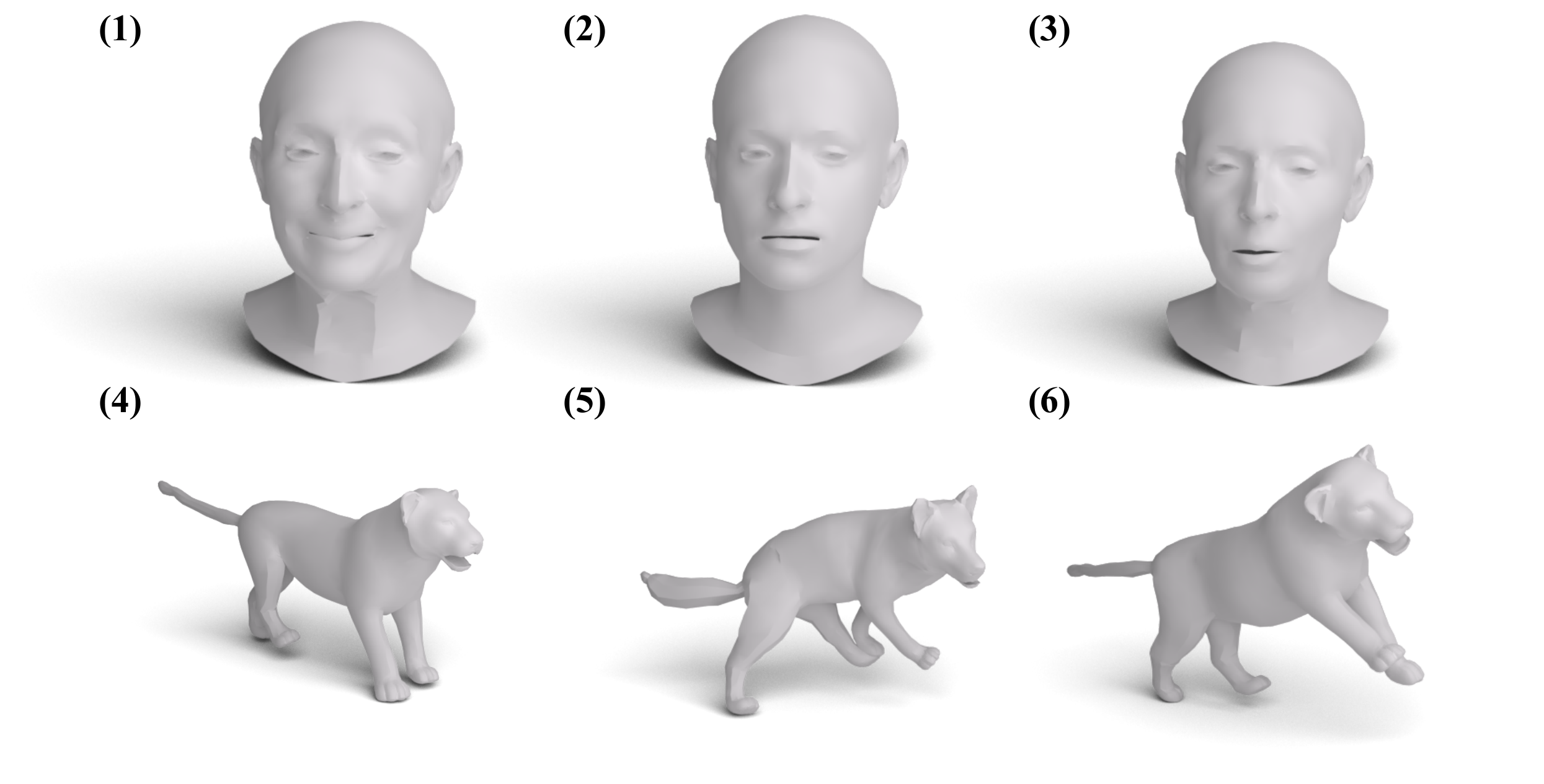}
\caption{{\bfseries \newtxt{The AE's reconstructions of the attack-detection quiz.}} \newtxt{The figure shows the AE's reconstructions of the shapes from the attack detection quiz (presented in Figure~\ref{fig:quiz_main}). The adversarial human faces were reconstructed by the AE as different people. The adversarial cow and the adversarial horse were transformed to leopards after passing through the AE. The clean shapes kept their input geometry.}}
\label{fig:quiz_recon}
\end{center} 
\end{figure}

%% file: supplementary/figures/stability/coma_stability_pdf.tex
\begin{figure}[tb!]
\begin{center}
\includegraphics[width=\columnwidth]{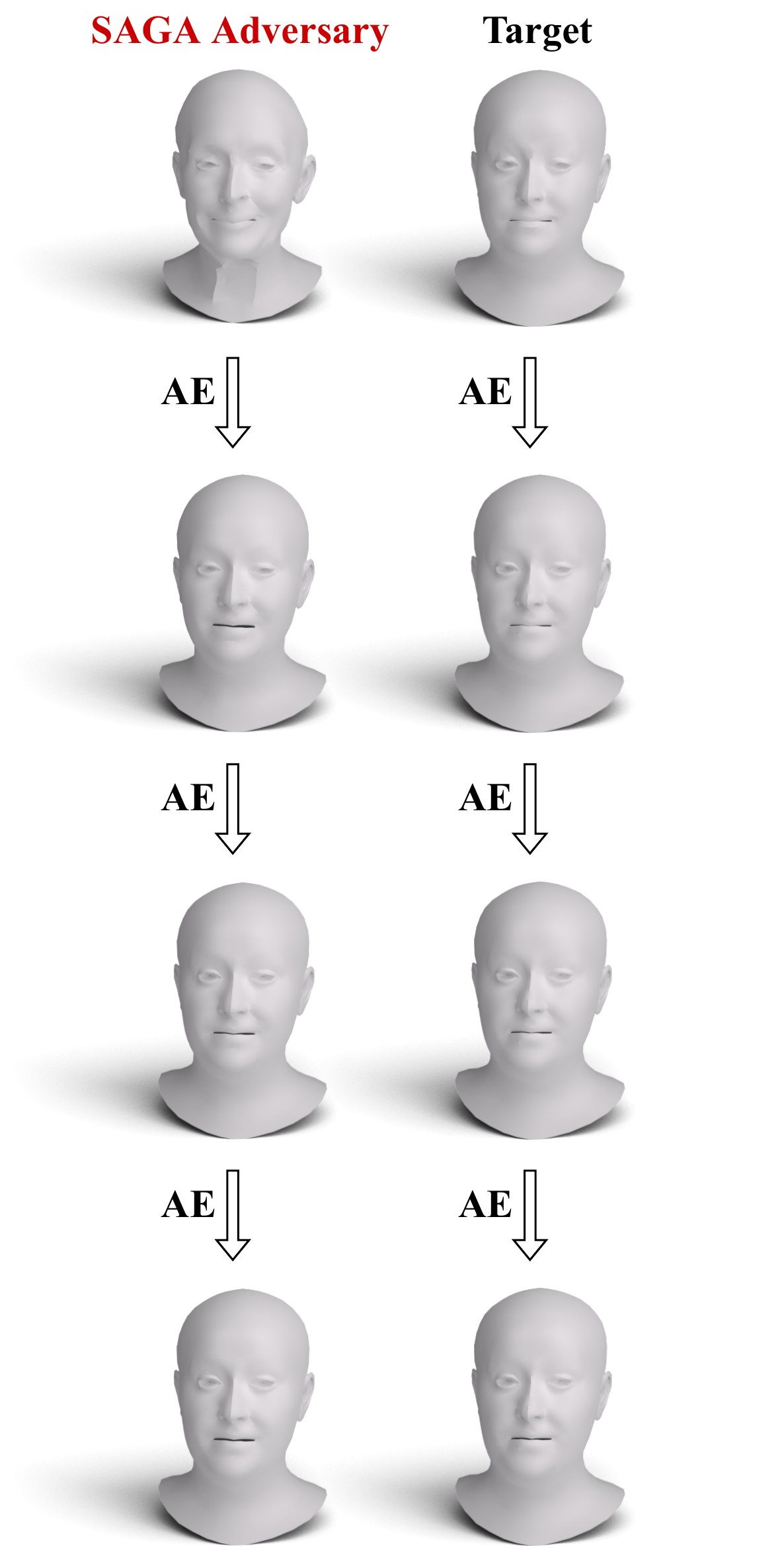}
\caption{{\bfseries Stability of the adversarial reconstructions of human faces.} The top row presents SAGA's adversarial example (left) and its clean target mesh (right). Each row below shows the reconstructions of the shapes from the row above it after passing through the AE. The shapes are taken from the CoMA dataset~\cite{ranjan2018generating}. Note that the source has a different identity than the target, with sharper facial features. The adversarial reconstruction keeps the target's geometry even after it is repeatedly encoded and reconstructed, and it is very similar to the reconstructions of the clean target shape.}
\label{fig:coma_stability}
\end{center} 
\end{figure}

%% file: supplementary/figures/stability/smal_stability_pdf.tex
\begin{figure}[tb!]
\begin{center}
\includegraphics[width=\columnwidth]{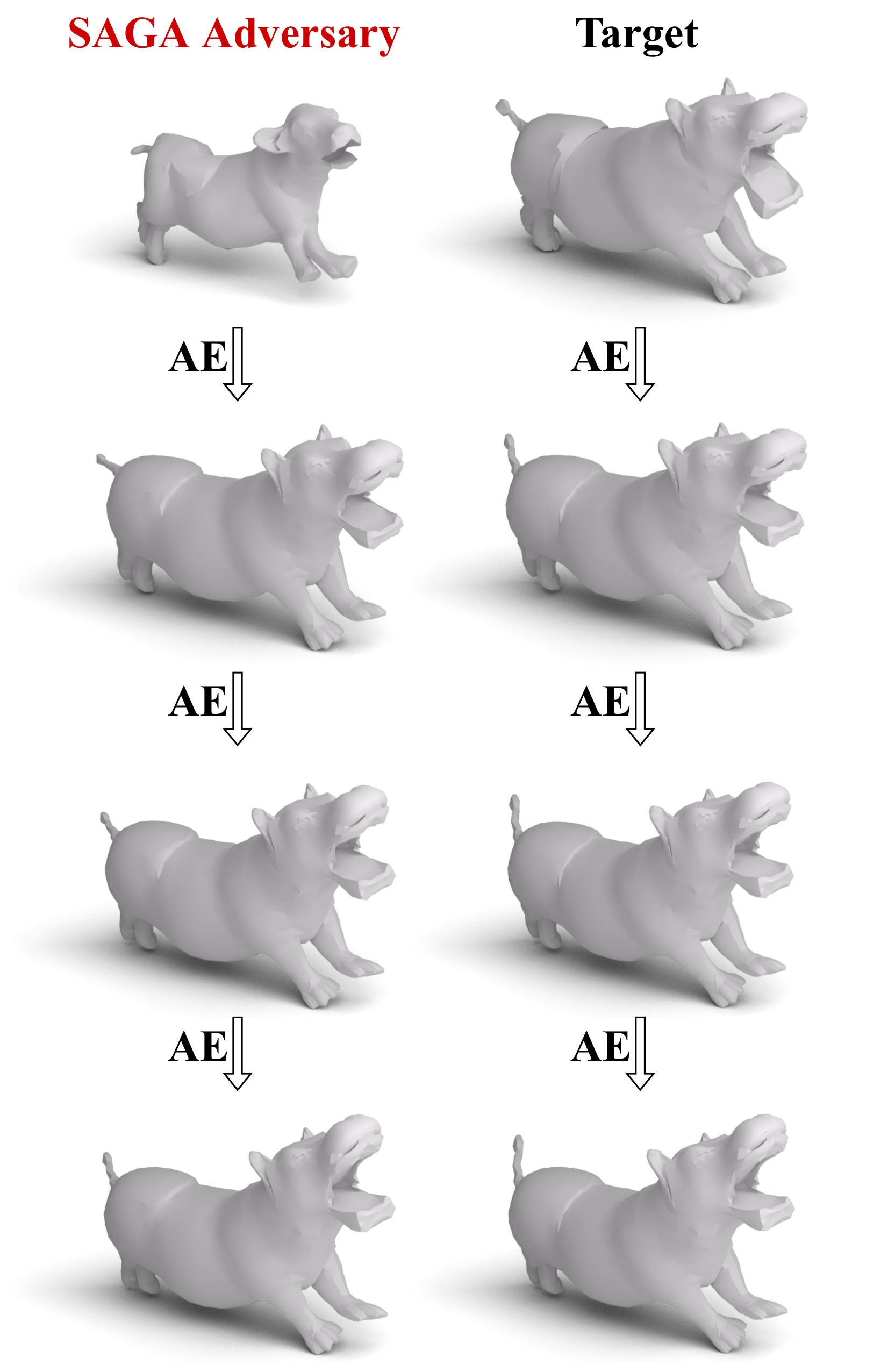}
\caption{{\bfseries Stability of the adversarial reconstructions of animals.} The top row presents SAGA's adversarial example (a \textit{cow}) and its clean target mesh (a \textit{hippo}) from left to right. Each row below shows the reconstructions of the shapes from the row above it after passing through the AE. The shapes are taken from the SMAL dataset~\cite{zuffi20173dmenagerie}. The adversarial reconstruction keeps the hippo's geometry even after it is repeatedly encoded and reconstructed, and it is very similar to the reconstructions of the clean target shape.}
\label{fig:smal_stability}
\end{center} 
\end{figure}

%% file: supplementary/figures/failures/coma_failure_pdf.tex
\begin{figure}[tb!]
\begin{center}
\includegraphics[width=\columnwidth]{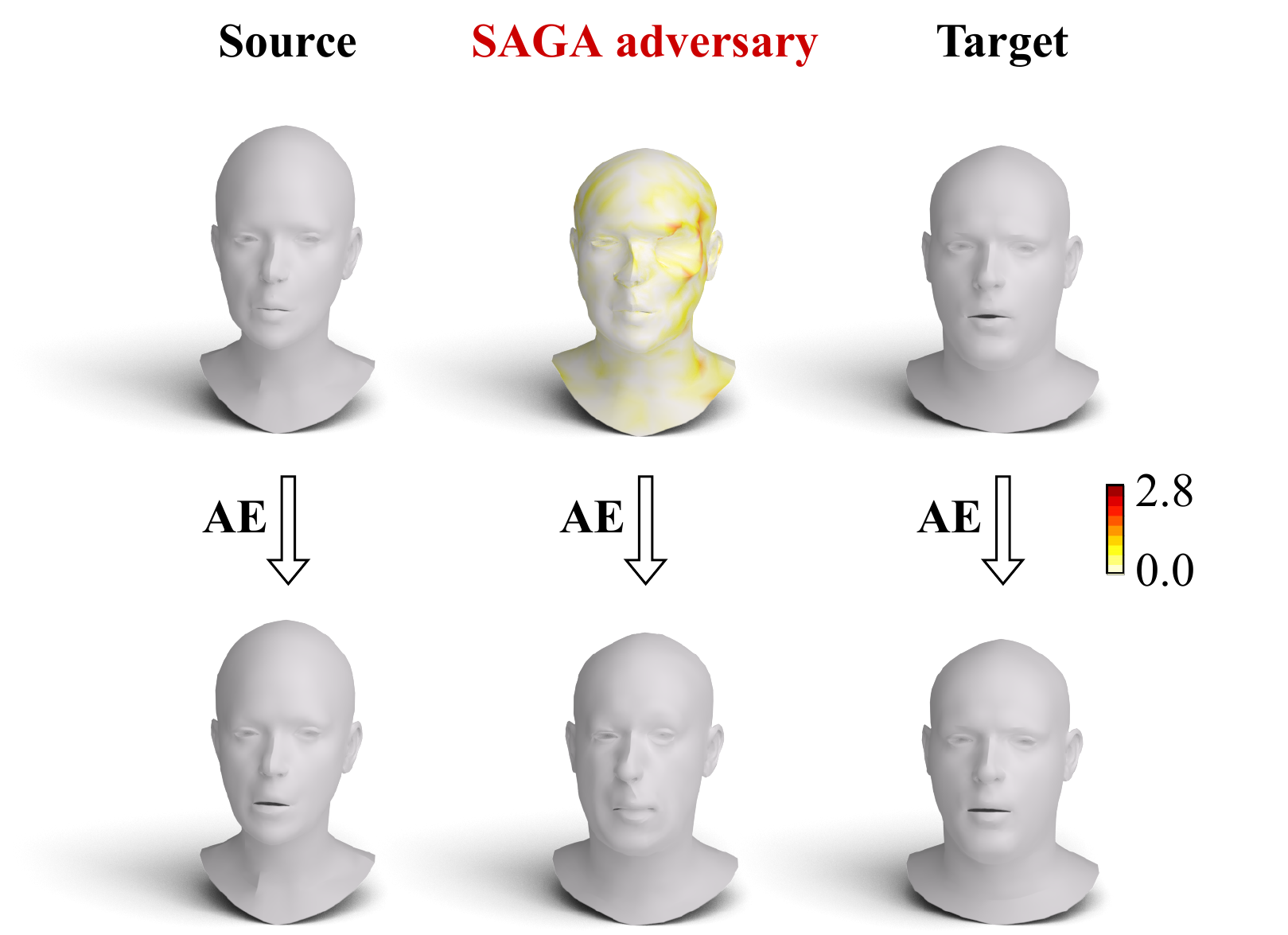}
\caption{{\bfseries A failure case on human faces.} A typical failure of SAGA on the CoMA dataset~\cite{ranjan2018generating}. Top row, left to right: the clean source mesh, SAGA's adversarial example, and the clean target shape. Bottom row: the reconstructions of the shapes from the top row after passing through the AE. The heatmap encodes the per-vertex curvature distortion values between the adversarial example and the clean source shape, growing from white to red. When the target geometry is too distant from the source, the perturbations change the source's identity. Also, the adversarial mesh and its reconstruction include visible deformations.}
\label{fig:coma_fail}
\end{center} 
\end{figure}

%% file: supplementary/figures/failures/smal_failure_pdf.tex
\begin{figure}[tb!]
\begin{center}
\includegraphics[width=\columnwidth]{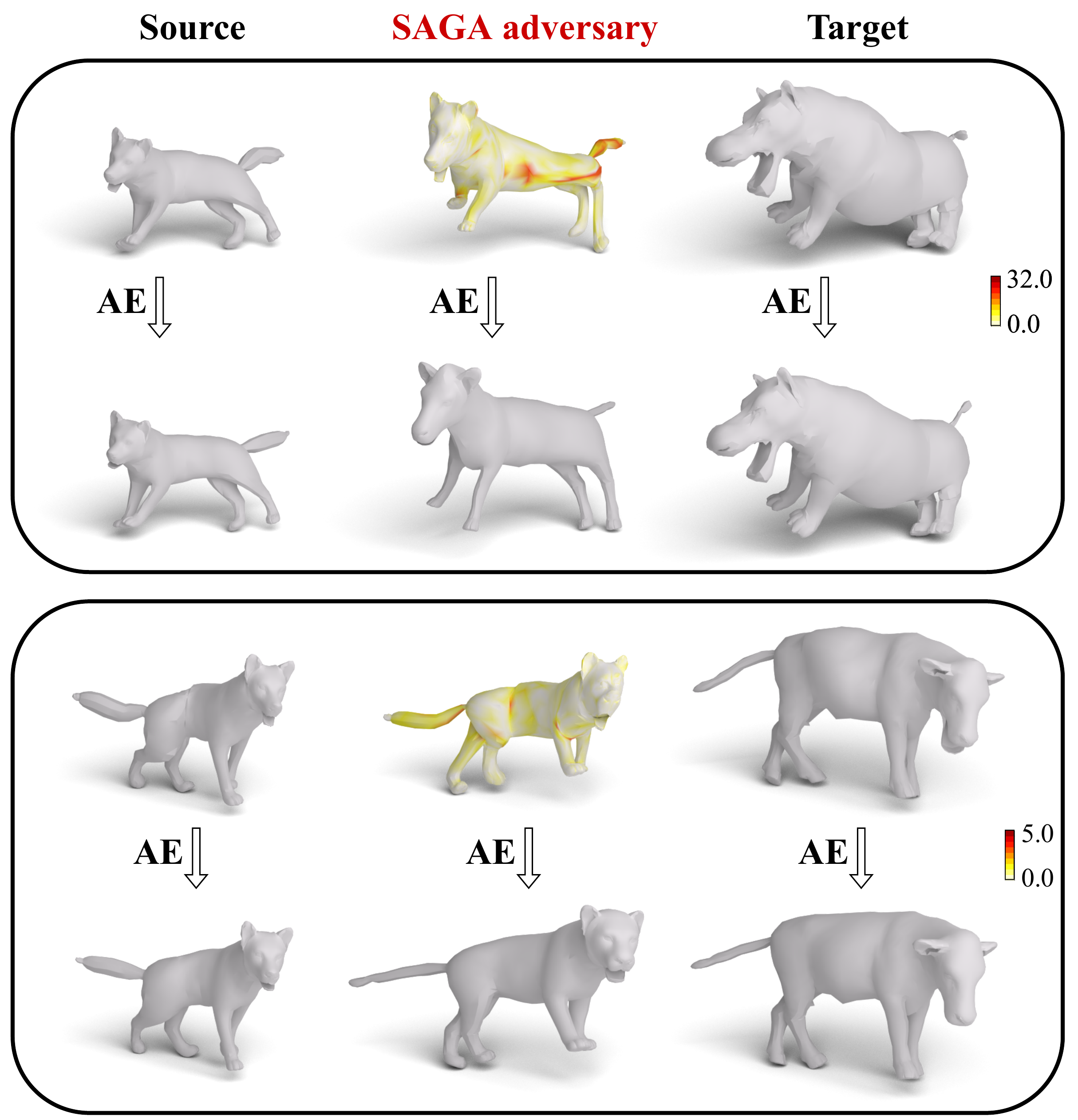}
\caption{{\bfseries Failure cases on animals.} Each frame shows a failure case of SAGA on the SMAL dataset~\cite{zuffi20173dmenagerie}. In each frame, top row, left to right: the clean source mesh (a \textit{dog}), SAGA's adversarial example, and the clean target shape. Bottom row: the reconstructions of the shapes from the top row after passing through the AE. The heatmap encodes the per-vertex curvature distortion values between the adversarial example and the clean source shape, growing from white to red. If the source and target shapes differ too much, the perturbations cause visible distortions to the source shape, and the reconstruction does not reach the desired target geometry. In the presented examples, the AE's reconstructions are of a \textit{horse} and a \textit{leopard} instead of the target geometries of a \textit{hippo} and a \textit{cow}.}
\label{fig:smal_fail}
\end{center} 
\end{figure}

%% file: supplementary/figures/evolution/coma_evolution_pdf.tex
\begin{figure*}[tb!]
\begin{center}
\includegraphics[width=\linewidth]{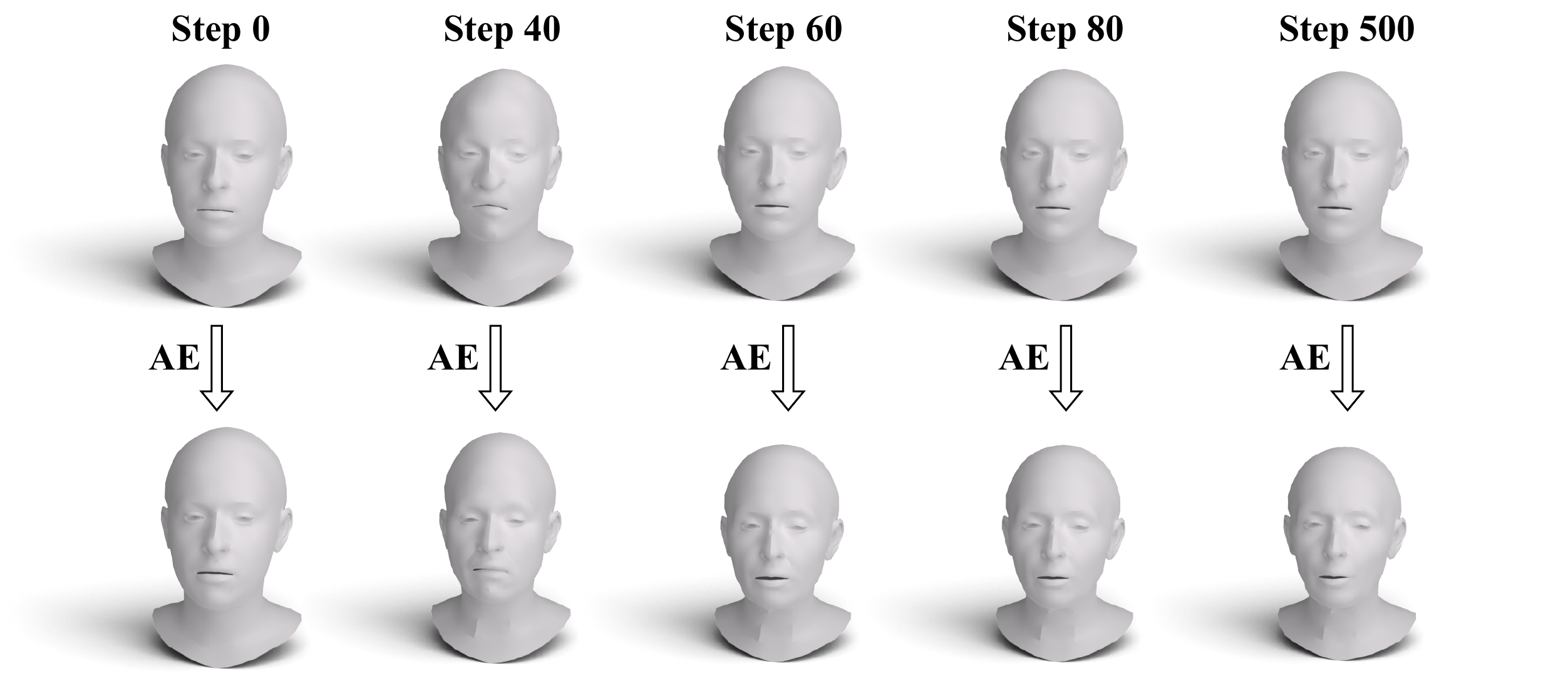}
\caption{{\bfseries Attack evolution on a human face.} Top row: an adversarial example from the CoMA dataset~\cite{ranjan2018generating} in different attack steps. Bottom row: the reconstructions of each adversarial shape by the AE. The leftmost column shows the initial state of the attack with no perturbation, and the rightmost column is the ending step of the attack. SAGA reaches a decent result quickly at gradient step $60$. The remaining attack steps fine-tune the result.}
\label{fig:coma_evolution}
\end{center} 
\end{figure*}

%% file: supplementary/figures/evolution/smal_evolution_pdf.tex
\begin{figure*}[tb!]
\begin{center}
\includegraphics[width=\linewidth]{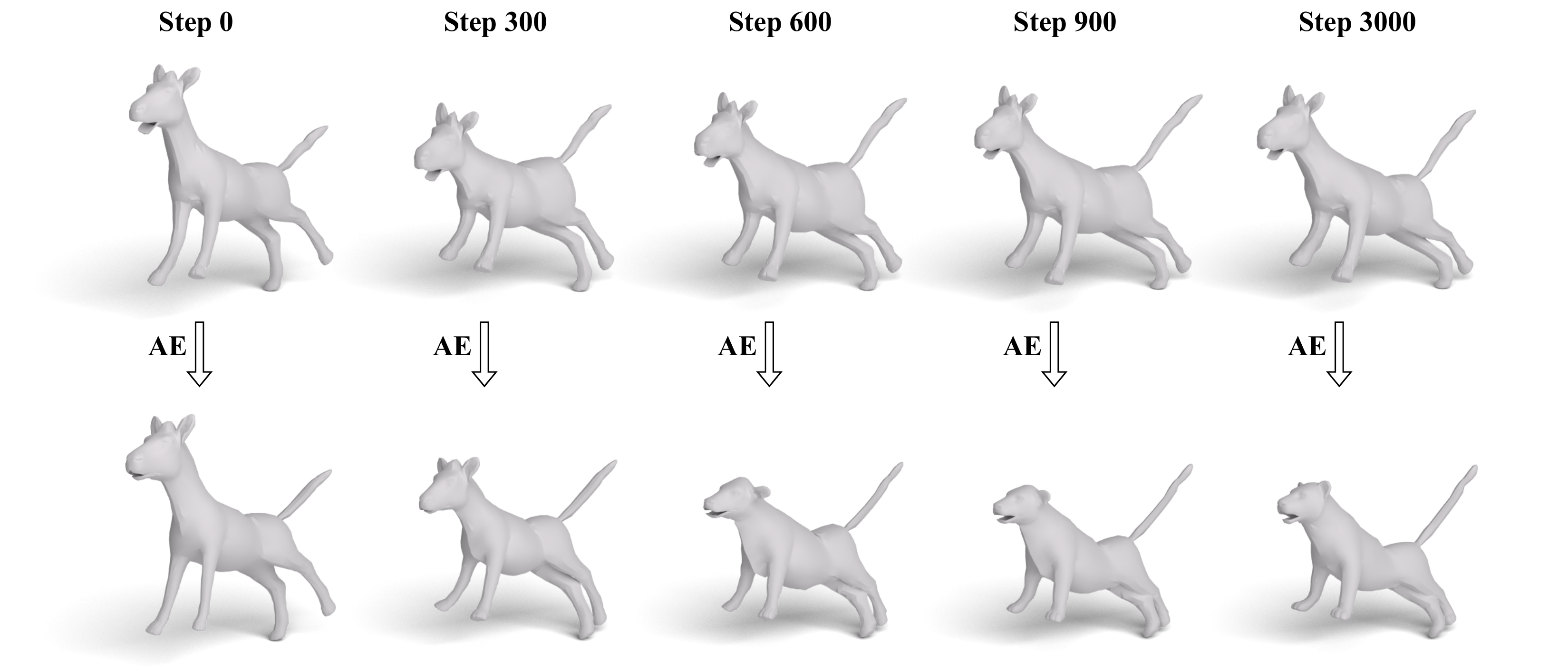}
\caption{{\bfseries Attack evolution on animals.} Top row: an adversarial example from the SMAL dataset~\cite{zuffi20173dmenagerie} in different attack steps. Bottom row: the reconstructions of each adversarial shape by the AE. The leftmost column shows the initial state of the attack with no perturbation, and the rightmost column is the ending step of the attack. The reconstructed shape smoothly turns from a \textit{horse} into a \textit{leopard} while changing to a \textit{cow} in between.}
\label{fig:smal_evolution}
\end{center} 
\end{figure*}

%% file: supplementary/figures/attack/coma_concat_1_pdf.tex
\begin{figure*}[tb!]
\begin{center}
\includegraphics[width=\textwidth,height=0.9\textheight,keepaspectratio]{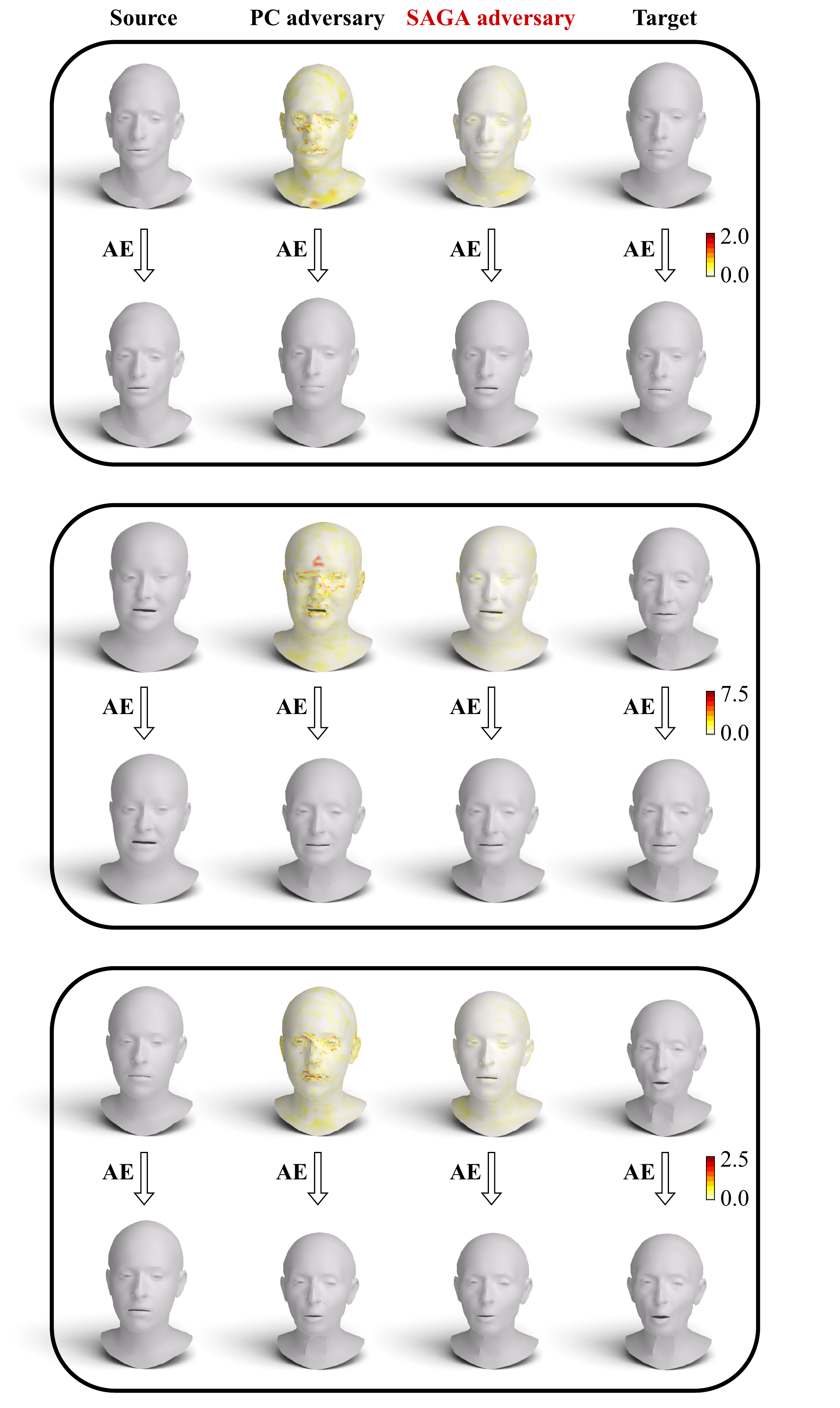}
\caption{{\bfseries Geometric attacks comparison.} The results of our attack are compared to Lang \etal's point cloud (PC) attack~\cite{lang2021geometric} on the CoMA dataset~\cite{ranjan2018generating}. Each frame presents a different source-target pair. In each frame, top row, left to right: the clean source mesh, the PC attack's adversarial example, SAGA's adversarial example, and the clean target shape. Bottom row: the reconstructions of the shapes from the top row after passing through the AE. The heatmap encodes the per-vertex curvature distortion values between each adversarial example and the clean source shape, growing from white to red. Our attack crafts adversarial examples with less distortion.}
\label{fig:coma_concat_1}
\end{center} 
\end{figure*}

%% file: supplementary/figures/attack/coma_concat_2_pdf.tex
\begin{figure*}[tb!]
\begin{center}
\includegraphics[width=\textwidth,height=0.9\textheight,keepaspectratio]{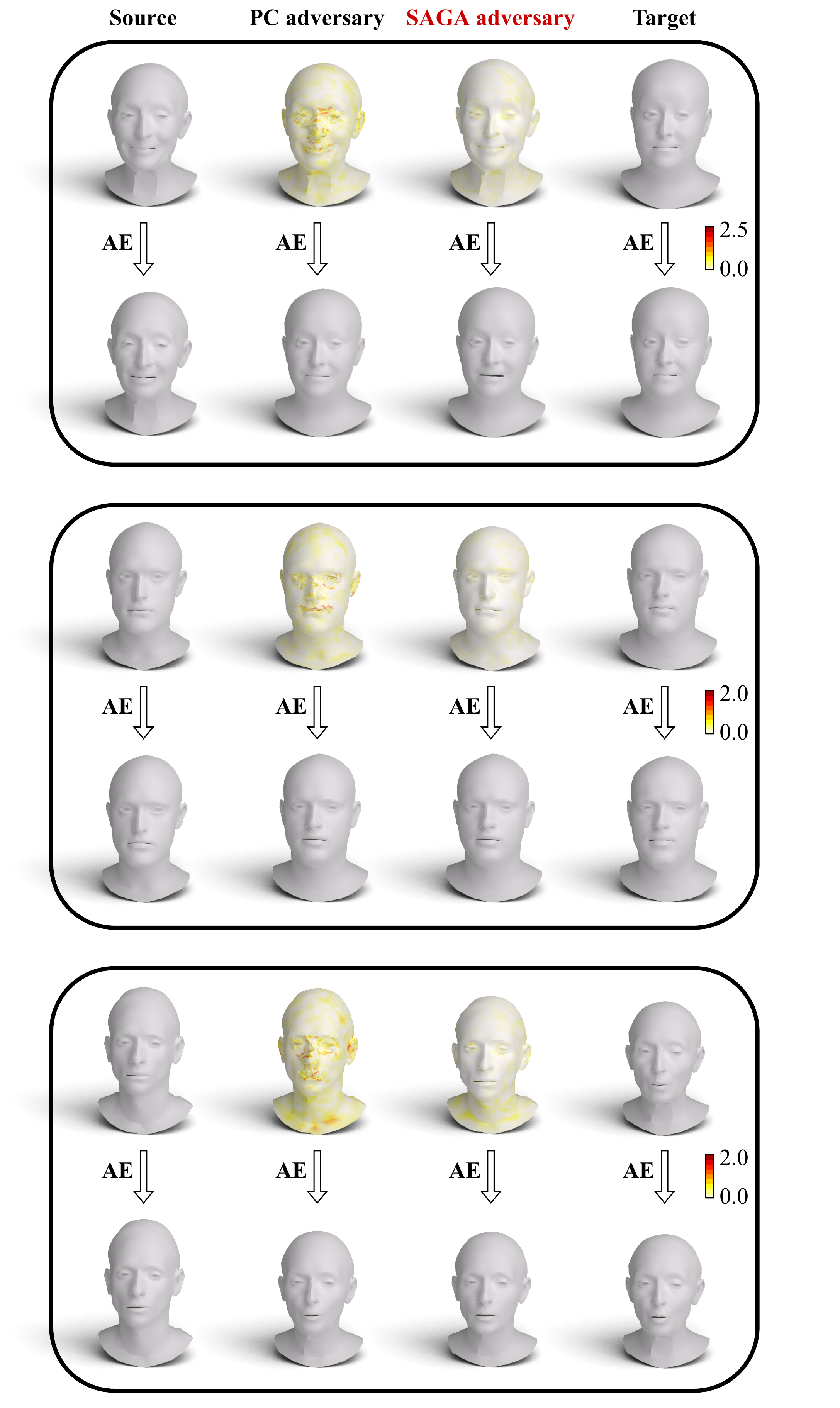}
\caption{{\bfseries Geometric attacks comparison.} Additional results on human faces, as described in Figure~\ref{fig:coma_concat_1}.}
\label{fig:coma_concat_2}
\end{center} 
\end{figure*}

%% file: supplementary/figures/attack/smal_concat_1_pdf.tex
\begin{figure*}[tb!]
\begin{center}
\includegraphics[width=0.96\textwidth,height=0.92\textheight,keepaspectratio]{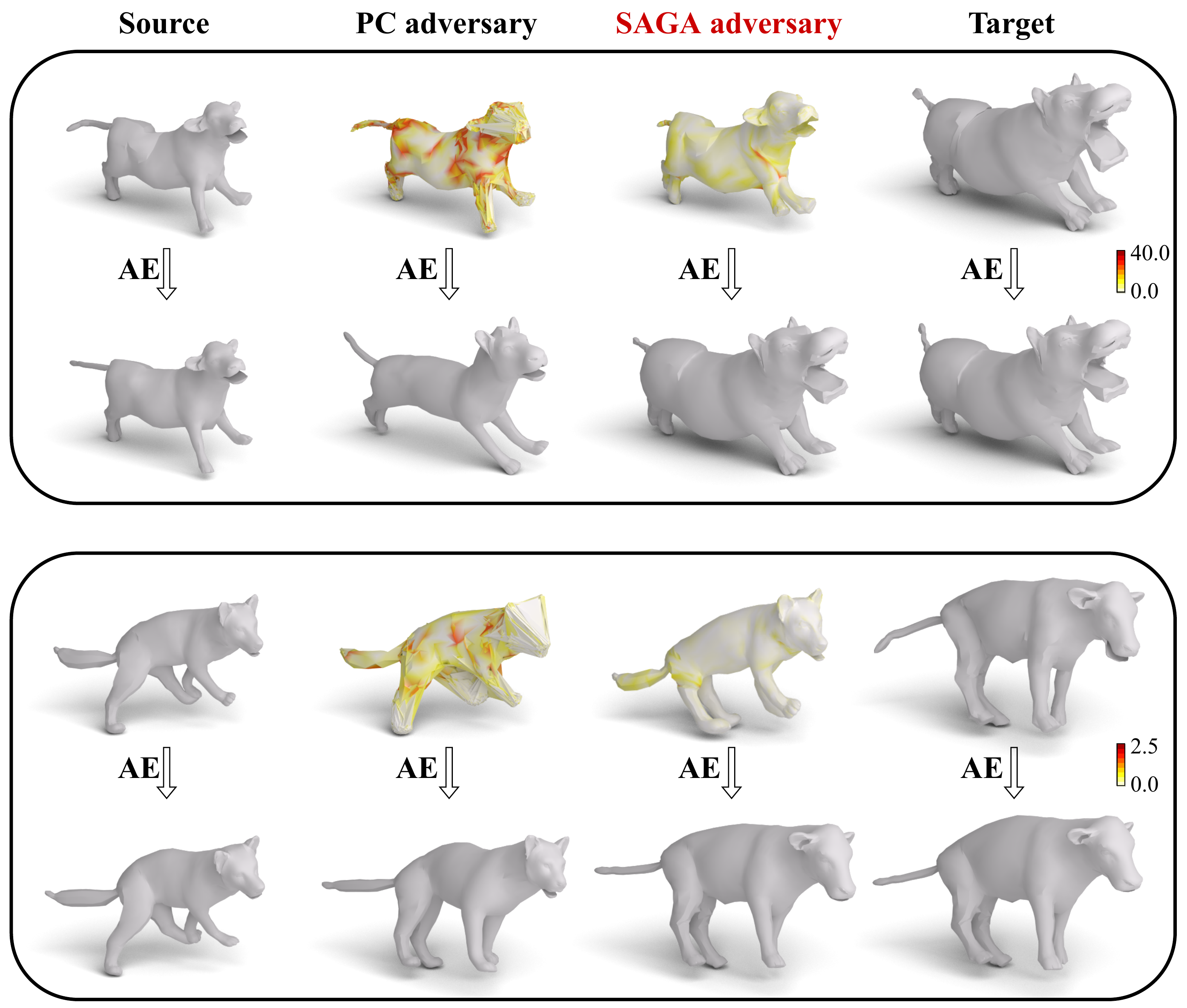}
\caption{{\bfseries Geometric attacks comparison.} The results of our attack are compared to Lang \etal's point cloud (PC) attack~\cite{lang2021geometric} on the SMAL dataset~\cite{zuffi20173dmenagerie}. Each frame presents a different source-target pair. In each frame, top row, left to right: the clean source mesh, the PC attack's adversarial example, SAGA's adversarial example, and the clean target shape. Bottom row: the reconstructions of the shapes from the top row after passing through the AE. The heatmap encodes the per-vertex curvature distortion values between each adversarial example and the clean source shape, growing from white to red. Lang \etal's attack severely distorts the source and does not reconstruct the desired target geometry. In contrast, our SAGA better preserves the source shape and achieves the target reconstruction.}
\label{fig:smal_concat_1}
\end{center} 
\end{figure*}

%% file: supplementary/figures/attack/smal_concat_2_pdf.tex
\begin{figure*}[tb!]
\begin{center}
\includegraphics[width=\textwidth,height=0.97\textheight,keepaspectratio]{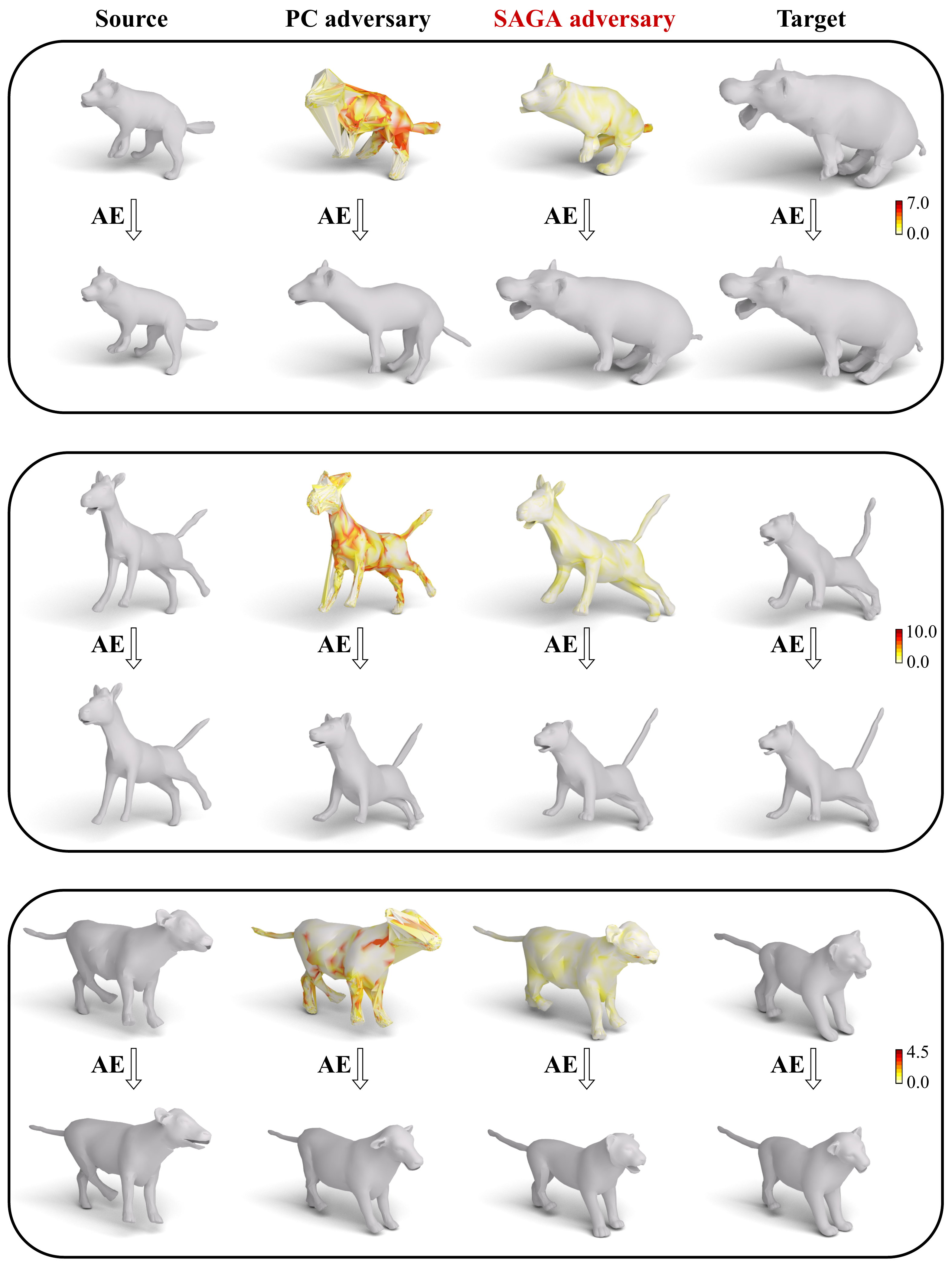}
%\vspace{-10pt}  % to narrow the gap between the fig and caption. you may increase/decrease the -10pt, but make sure that is does not look like you play with the paper template too much...
%\vspace{-10pt}  % to narrow the gap between the caption and the paper text below
\caption{{\bfseries Geometric attacks comparison.} Additional results on animals, as described in Figure~\ref{fig:smal_concat_1}.}
\label{fig:smal_concat_2}
\end{center} 
\end{figure*}